\documentclass[10pt,twocolumn,letterpaper]{article}

\usepackage{cvpr}              %

\usepackage[dvipsnames]{xcolor}

\usepackage{graphicx}
\usepackage{amsmath}
\usepackage{amssymb}

\usepackage{siunitx}
\usepackage{booktabs,makecell}
\usepackage{etoolbox}
\usepackage{color, colortbl}

\usepackage[ruled,vlined]{algorithm2e}

\usepackage{setspace}

\usepackage[title]{appendix}

\usepackage{enumitem}
\setlist{topsep=0pt, leftmargin=*,noitemsep,topsep=0pt,parsep=0pt,partopsep=0pt}

\usepackage[T1]{fontenc}
\usepackage{acronym}
\usepackage{caption}
\usepackage{subcaption}
\usepackage{cuted}
\usepackage{adjustbox}
\usepackage{pifont}
\usepackage{algpseudocode}
\usepackage{multirow}
\usepackage{tablefootnote}

\newcommand{\cmark}{\ding{51}}%
\newcommand{\xmark}{\ding{55}}%

\acrodef{bts}[BTS]{Behind the Scenes}
\acrodef{mde}[MDE]{monocular depth estimation}
\acrodef{ssc}[SSC]{semantic scene completion}
\acrodef{mvbts}[MVBTS]{Multi-View Behind the Scenes}
\acrodef{kdbts}[KDBTS]{Knowledge-Distillation Behind the Scenes}
\acrodef{nerf}[NeRF]{neural radiance field}
\acrodef{nvs}[NVS]{novel view synthesis}
\acrodef{mlp}[MLP]{multi-layer perceptron}

\newcommand{\pseudoparagraph}[1]{\textbf{#1}}

\definecolor{Gray}{gray}{0.96}

\definecolor{cvprblue}{rgb}{0.21,0.49,0.74}
\usepackage[pagebackref,breaklinks,colorlinks,citecolor=cvprblue]{hyperref}

\usepackage[capitalize]{cleveref}
\crefname{section}{Sec.}{Secs.}
\Crefname{section}{Section}{Sections}
\Crefname{table}{Table}{Tables}
\crefname{table}{Tab.}{Tabs.}

\def\paperID{2805} %
\def\confName{CVPR}
\def\confYear{2024}

\title{Boosting Self-Supervision for Single-View Scene Completion \\ via Knowledge Distillation}

\author{Keonhee Han$^{\text{*},1}$ \hspace{1cm} Dominik Muhle$^{\text{*},1,2}$ \hspace{1cm} Felix Wimbauer$^{1,2}$ \hspace{1cm} Daniel Cremers$^{1,2}$\\
$^1$TU Munich \hspace{1cm} $^2$Munich Center for Machine Learning \hspace{1cm} 
\\
{\tt\small \{keonhee.han, dominik.muhle, felix.wimbauer, cremers\}@tum.de}
}

\newcommand\blfootnote[1]{%
  \begingroup
  \renewcommand\thefootnote{}\footnote{#1}%
  \addtocounter{footnote}{-1}%
  \endgroup
}

\begin{document}
\maketitle
\begin{strip}
\vspace{-1.6cm}
\centering
\captionsetup{type=figure}
\includegraphics[trim={2cm 0cm 0cm 0cm},clip,width=\linewidth]{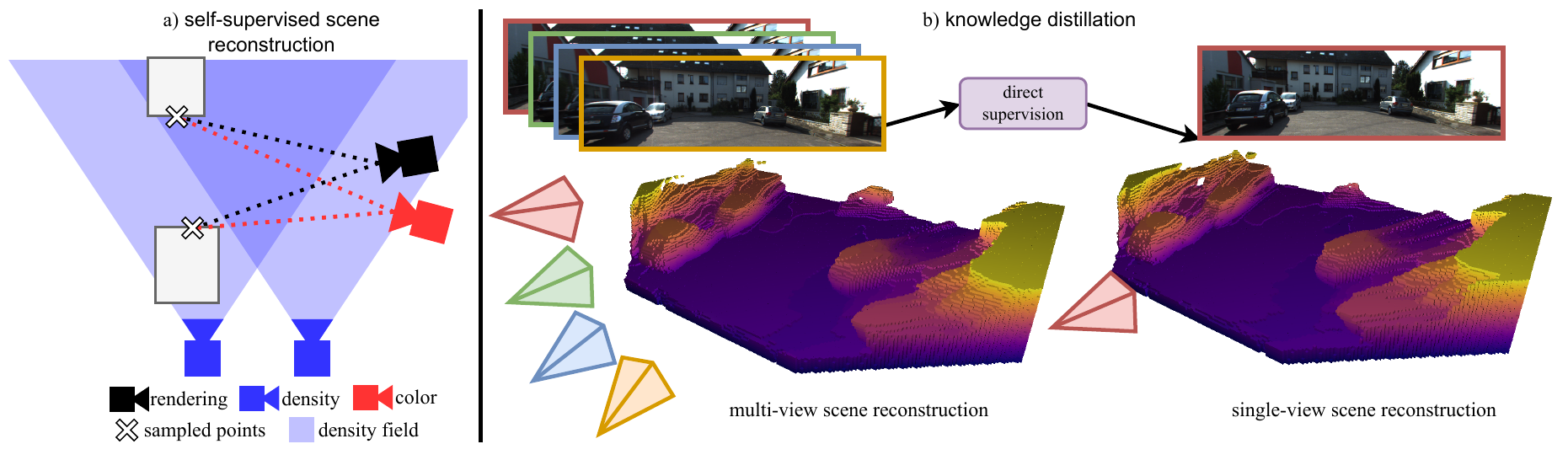}
\captionof{figure}{
    \textbf{Knowledge Distillation from Multi-View to Single-View.} We propose to boost single-view scene completion by exploiting additional information from multiple images. a) we first train a novel multi-view scene reconstruction algorithm that is able to fuse density fields from multiple images in a fully self-supervised manner. b) we then employ knowledge distillation to directly supervise a state-of-the-art single-view reconstruction model in 3D to boost its performance.  \href{https://keonhee-han.github.io/publications/kdbts/}{https://keonhee-han.github.io/publications/kdbts/}
}
\label{fig:teaser}
\end{strip}

\begin{abstract}
Inferring scene geometry from images via Structure from Motion is a long-standing and fundamental problem in computer vision. While classical approaches and, more recently, depth map predictions only focus on the visible parts of a scene, the task of scene completion aims to reason about geometry even in occluded regions. With the popularity of \acp{nerf}, implicit representations also became popular for scene completion by predicting so-called density fields. Unlike explicit approaches \eg voxel-based methods, density fields also allow for accurate depth prediction and novel-view synthesis via image-based rendering.
In this work, we propose to fuse the scene reconstruction from multiple images and distill this knowledge into a more accurate single-view scene reconstruction. To this end, we propose \ac{mvbts} to fuse density fields from multiple posed images, trained fully self-supervised only from image data. 
Using knowledge distillation, we use \ac{mvbts} to train a single-view scene completion network via direct supervision called KDBTS. It achieves state-of-the-art performance on occupancy prediction, especially in occluded regions.

\vspace{-.3cm}
\end{abstract}
  
\blfootnote{$^{\text{*}}$ equal contribution.} %
\vspace{-1.65cm}
\section{Introduction}
\label{sec:intro}

Obtaining accurate 3D geometry of a scene is crucial to achieving a holistic understanding of our surroundings. This knowledge enables a plethora of tasks both in robotics and autonomous driving. Reconstructing the scene's geometry from raw images, also called Structure-from-Motion \cite{schoenberger2016sfm}, is a long-standing task in computer vision. Classical methods of scene reconstruction rely on multiple images and triangulation of matched keypoints,  such as Harris Corner Detection, SIFT, and ORB \cite{Harris88alvey, SIFT_Lowe, ORB_conf}, to infer the geometry of a scene. More recently, the breakthrough of deep learning gave rise to \ac{mde} methods \cite{godard2017unsupervised, godard2019digging}, that allowed to predict a dense per pixel depth from only a single image. These are either trained with ground truth supervision or in a self-supervised manner from video data \cite{zhou2017unsupervised, zhan2018unsupervised, godard2019digging, luo2019every, guizilini20203d, shu2020feature, yuan2022new, gonzalezbello2020forget, watson2019self}. Despite the impressive results of \ac{mde} for geometry estimation, the depth map representations are limited in their expressiveness. Projecting them into other frames will lead to holes or possibly wrong estimates due to occlusions in the scene. Furthermore, they cannot make any statements about the scene beyond the visible surfaces. 

Scene reconstruction broadens the task of scene representation from depth maps to reasoning about the whole geometry of a scene, including occluded regions \cite{song2017semantic} by doing shape completion. \cite{song2017semantic} introduced the task of semantic scene completion, which extends scene reconstruction to include semantic predictions. The 3D nature of both tasks results in many methods relying on voxel representations for their predictions that either use images \cite{cao2022monoscene, li2023voxformer} or LiDAR/depth maps \cite{roldao2020lmscnet, song2017semantic} as input. 

Neural implicit representations in the form of neural fields offer an alternative scene representation. Popularized by \acp{nerf}, which models a scene's geometry and view-dependent colors in the weights of a neural network, they allow for photorealistic novel view synthesis. However, the original \ac{nerf} work \cite{mildenhall2021nerf} cannot generalize to different scenes and often suffers from inaccurate geometry \cite{verbin2022ref}. These limitations were addressed by many follow-up works to \ac{nerf} \cite{yu2021pixelnerf, niemeyer2022regnerf, verbin2022ref}. Similarly, density fields \cite{wimbauer2023behind} solely model the scene's geometry, addressing the geometric inconsistencies of \acp{nerf}. The work \ac{bts} \cite{wimbauer2023behind} uses density fields predicted from a single image for scene completion, while IBRnet and NeuRay leverage density predictions from multiple images together with images-based rendering for novel-view synthesis.
\ac{mvbts} presents an extension of \ac{bts} from a single image to multiple images for more accurate scene reconstructions. In the absence of accurate 3D ground truth geometry, we leverage these scene reconstructions to directly supervise single-view density prediction models. We retrain the original \ac{bts} architecture in this pseudo-supervised setting as KDBTS to boost its performance.
We release the code to further facilitate research.

Our \textbf{contributions} within this work are threefold:
\begin{itemize}
    \item We extend the density field prediction of \ac{bts} to a multi-view setting trained in a fully self-supervised manner.
    \item We propose knowledge distillation training to directly supervise single-view density fields on fused multi-view predictions.
    \item Our method achieves state-of-the-art performance for both multi- and single-view occupancy prediction on the KITTI-360 benchmark.
\end{itemize}

\section{Related Work}
\label{sec:related_work}

\pseudoparagraph{Neural Scene Representations}
\Acf{nerf} \cite{mildenhall2021nerf} has popularized the method of representing a scene with neural implicit functions. \Ac{nerf} allows to query color and occupancy of a scene in continuous space by storing the scene within the weights of a neural network. This allows \ac{nerf} to do photorealistic \ac{nvs} of a scene. A downside of the original \ac{nerf} is the need to retrain the network weights for each scene. This was addressed by works such as PixelNeRF \cite{yu2021pixelnerf} and MINE \cite{li2021mine} that construct neural radiance fields on the fly conditioned on sparse input views. \cite{sharma2023neural} maps 2D image observations of a scene to a persistent 3D scene representation, disentangling movable and immovable components of the scene. Neo360 \cite{irshad2023neo} uses a tri-planar representation together with foreground and background MLPs to allow for representing unbounded scenes.

However, most \ac{nerf} methods focus on tasks such as novel view synthesis, often leading to inaccurately reconstructed scene geometry. Works such as \cite{verbin2022ref} address this by modeling effects such as reflections on surfaces.

\pseudoparagraph{Image Based Rendering}
Imaged-based rendering synthesizes images for novel views based on the color of surrounding reference images. Early methods focused on blending reference pixels based on geometry proxies \cite{buehler2001unstructured, debevec1996modeling, heigl1999plenoptic}. Other works on light fields reconstruct 4D plenoptic functions from input views \cite{gortler1996lumigraph, davis2012unstructured}. More recent approaches are based on \ac{nerf} architectures. IBRnet \cite{wang2021ibrnet} uses input images to construct a radiance field and density field on the fly. The density is predicted from pooled feature vectors in a ray transformer, sharing information between 3D points. The color is reconstructed as a weighted sum of the colors of surrounding views. GeoNeRF \cite{johari2022geonerf} uses cost volumes together with an attention-based aggregation for density prediction while replacing the ray transformer with an autoencoder. NeuRay \cite{liu2022neuray} addresses the problem of inconsistent image features due to occlusions. Based on either cost volumes or depth maps, it predicts the visibility of 3D points in different views, leading to better renderings. 

\pseudoparagraph{Scene Completion}
Scene completion, sometimes also called scene reconstruction, refers to the task of estimating the 3D geometry of a scene from images. Depending on the task, scene completion can range from depth prediction \cite{godard2017unsupervised, godard2019digging, spencer2020defeat, lyu2021hr, wimbauer2021monorec} to reconstructing even occluded parts of a scene \cite{wimbauer2023behind, cao2022monoscene}. Depth prediction can either come from monocular images \cite{godard2019digging, zhou2021diffnet} or stereo images \cite{zhang2019ga, Khamis_2018_stereonet}. Especially \ac{mde} has been successfully used in monocular visual odometry applications \cite{yang2018deep, koestler2022tandem}.

The information from depth prediction is restricted to the visible surfaces of a scene, leaving out important information about the shapes of objects. Given a depth map \cite{firman2016structured, wu20153d, song2017semantic} complete shape geometry in an environment using voxel-grid representations. Voxlets \cite{firman2016structured} use random forests to predict signed distance values in the grid, whereas \cite{wu20153d} models a probabilistic distribution on voxels. Generalizable \ac{nerf} methods such as PixelNeRF \cite{yu2021pixelnerf} and MINE \cite{li2021mine} can also be used to complete the scene geometry. Recently, \acf{bts} \cite{wimbauer2023behind} introduced a density field representation based on \ac{nerf} to model the scene geometry implicitly together with imaged-based rendering to explicitly learn to ``look behind'' objects. 

\begin{figure*}[ht]
\centering
\includegraphics[trim={1.5cm 0.5cm 0cm 0.3cm},clip,width=0.92\linewidth]{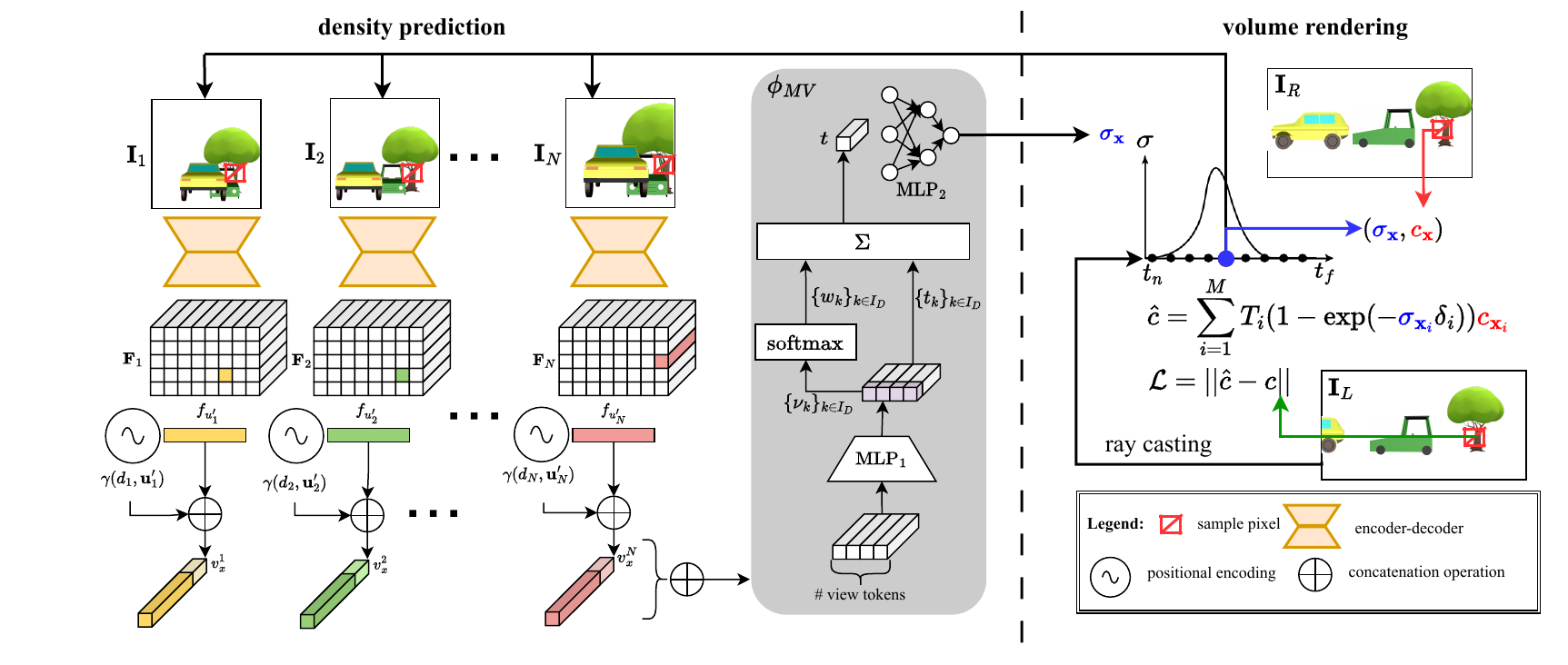}
\vspace{-0.2cm}
\caption{
\textbf{Overview.} Given multiple input images $\textbf{I}_k$ ($k \in I_D$) an encoder-decoder backbone predicts per image a pixel-aligned feature map $\textbf{F}_k$ (top left). The feature $f_{\textbf{u}}$ of pixel $\textbf{u}$ encodes the occupancy and confidence distribution of a ray cast through pixel $\textbf{u}$. Given a 3D point $\textbf{x}$ and its projections $\textbf{u}^\prime_k$ into the different camera images, we extract the corresponding feature vectors and positional embeddings $\gamma(d, \textbf{u})$. A multi-view network $\phi_\text{MV}$ decodes all feature vectors into a density prediction $\sigma_\textbf{x}$ (middle). Together with color samples from another image ($\textbf{I}_R$), this can be used to render novel views in an image-based rendering pipeline. 
$\textbf{I}_R$ is not required to be close to the input images, as our method can predict density in occluded regions.
See \autoref{fig:training_setup} for more details about the importance of covering the whole scene. We train our networks by using a photometric consistency loss of an image $\textbf{I}_L$ close to $\textbf{I}_R$ (right).
}
\label{fig:architecture}
\vspace{-0.6cm}
\end{figure*}

\Ac{ssc}, first introduced in \cite{song2017semantic}, extends this task even further to the semantic setting. For occupied regions, semantic labels are predicted. In the beginning, \ac{ssc} methods either focused on indoor scenarios with image data \cite{zhang2018efficient,  zhang2019cascaded, liu2018see, li2019depth, li2019rgbd, li2020anisotropic, chen20203d, cai2021semantic} or outdoor scenarios with LiDAR data \cite{roldao2020lmscnet, cheng2021s3cnet, yan2021sparse, li2021semisupervised, rist2021semantic}. MonoScene \cite{cao2022monoscene} was the first method to tackle outdoor scenarios from image data by using line-of-sight projection of deep image features. VoxFormer \cite{li2023voxformer} utilizes a transformer architecture with deformable attention layers. Both rely on voxels for their scene representation. S4C \cite{hayler2023s4c} uses an implicit scene representation and a self-supervised training approach to remove the need for annotated 3D data. 

In this work, we build upon the ideas of scene completion methods from single images, \ac{bts} \cite{wimbauer2023behind}. Specifically, to exploit all the available image data at inference time for more accurate occupancy predictions. We then leverage our accurate predictions to further improve single-image scene completion. This is in contrast to the works of \cite{wang2021ibrnet, liu2022neuray, johari2022geonerf} that focus on \ac{nvs} and rely on close-by views to provide both color and density information.

\section{Method}
\label{sec:method}
In the following, we briefly recap the \acf{bts} model as we extend its architecture and use it in our knowledge distillation scheme. We also cover training setup to predict the scene's geometry, even in occluded areas. We then introduce the extension of the \ac{bts} architecture to a multi-view setting and propose a knowledge distillation scheme for improved single-view predictions.

\pseudoparagraph{Notation}
We denote images by $\textbf{I}_{k} \in [0, 1]^{3\times H\times W} = (\mathbb{R}^3)^\Omega$ which are defined on a lattice $\Omega = \{1, \ldots, H\}\times\{1, \ldots, W\}$. The corresponding camera poses and projection matrices of the images are given as $T_i \in \mathbb{R}^{4\times 4}$ and $K_i \in \mathbb{R}^{3 \times 4}$, respectively. The density of a point $\textbf{x} \in \mathbb{R}^3$ in world coordinates is given by $\sigma_{\textbf{x}}$. Its projection into an image $k$ is given by $\pi_k(\textbf{x}) = K_k T_k \textbf{x}$ in homogeneous coordinates. 
From our set of all image indices $I = \{1,\ldots,N\}$, we create the following subsets $I_D$, $I_L$, and $I_R$ for density, loss, and render. For these subsets, the following holds:
\begin{align}
    & I_D \subset I \\
    & I_L \cup I_R = I \\
    & I_L \cap I_R = \emptyset
\end{align}
We give examples together with a more detailed explanation for these splits in the supplementary material.

\subsection{Prerequisites: Density Field from a Single View}
\ac{bts} models a scene by predicting a volumetric density for a 3D point $\textbf{x}$ conditioned on an input image $\textbf{I}_\text{I}$ \cite{wimbauer2023behind}. This is done by predicting a pixel-aligned feature map $\textbf{F}_\text{I} \in (\mathbb{R}^C)^\Omega$ with an encoder-decoder architecture from the image. The point $\textbf{x}$ is projected into the input image $\textbf{u}^\prime_\text{I} = \pi_{\text{I}}(\textbf{x})$ to obtain the corresponding feature vector $f_{\textbf{u}^\prime} = \textbf{F}(\textbf{u}^\prime)$ from the feature map. A small \ac{mlp} $\phi_{SV}$ decodes the density from the feature vector as well a positional encoding $\gamma(d_\text{I}, \textbf{u}^\prime_\text{I})$ of the depth $d_\text{I}$ and pixel position $\textbf{u}^\prime_\text{I}$ as
\begin{equation}
    \sigma_\textbf{x} = \phi_{SV}(f_{\textbf{u}^\prime_\text{I}}, \gamma(d_\text{I}, \textbf{u}^\prime_\text{I}))
\end{equation}
The intuition behind this density prediction is that the feature vector of a pixel learns to describe the density distribution along the corresponding ray \cite{wimbauer2023behind}. Together with the positional encodings, the MLP ($\phi_{SV}$) decodes this distribution into a density prediction. 

\subsection{MVBTS: Density Field from Multiple Views}
We extend the single-view density prediction by replacing the $\phi_{SV}$ with a confidence-based multi-view architecture
\begin{equation}
    \sigma_\textbf{x} = \phi_{MV}(\{f_{\textbf{u}^\prime_k}, \gamma(d_k, \textbf{u}^\prime_k)\}_{k\in I_D})\,,
\end{equation}
that is able to aggregate the information from the $|I_D|$ images. 
In the following, we will break down how to predict a density value for a point $\textbf{x}$ from multiple input images. 

For each of the input images  $\{\textbf{I}_{k}\}_{k \in I_{D}}$, we predict a pixel-aligned feature map $\textbf{F}_k$ with a shared encoder-decoder architecture. The point $\textbf{x}$ is then projected into each of the input images $\textbf{u}^\prime_k = \pi_{k}(\textbf{x})$, giving us $|I_D|$ feature vectors $f_{\textbf{u}^\prime_k} = \textbf{F}_k(\textbf{u}^\prime_k)$. We keep track of valid projections into the frustum of a camera with a binary masking vector $\textbf{m} \in \{0, 1\}^{|I_D|}$. \label{binary masking vector} The feature vectors are concatenated with the respective positional encodings $\gamma(d_k, \textbf{u}^\prime_k)$ and fed through a small \ac{mlp}
\begin{equation}
    (\nu_k, t_k) = \text{MLP}_1(f_{\textbf{u}^\prime_k}, \gamma(d_k, \textbf{u}^\prime_k))
\end{equation}
to produce both a confidence $\nu_k$ and a view-dependent features $t_k$. A softmax layer then converts the confidence values into weights
\begin{equation}
    \{w_k\}_{k \in I_{D}} = \text{softmax}(\{\nu_k\}_{k \in I_D}, \textbf{m}) \,,
\end{equation}
that are then used in a weighted summation 
\begin{equation}
    t = \sum_{k \in I_{D}} w_k t_k
\end{equation}
to produce a \textit{view-aggregated} feature $t$. This feature is then decoded by a second \ac{mlp} to produce the density estimate
\begin{equation}
    \sigma_\textbf{x} = \text{MLP}_2(t)\,.
\end{equation}
For more details, see the left-hand side of \autoref{fig:architecture}.

\begin{figure}
\centering
\captionsetup{type=figure}
\includegraphics[trim={0cm 0cm 0cm 0cm},clip,width=\linewidth]{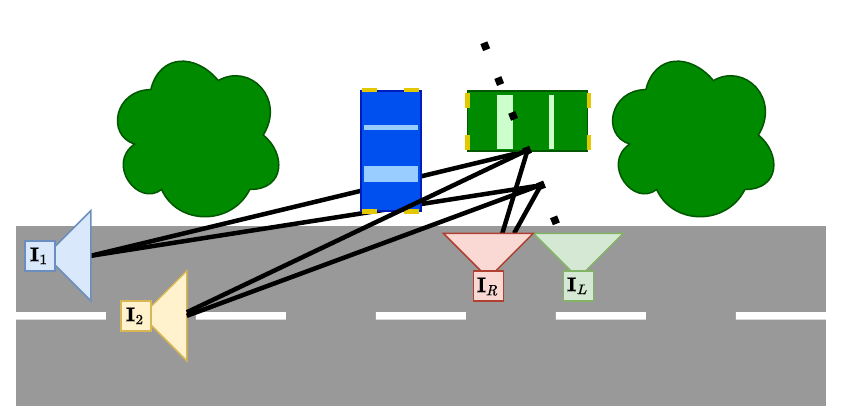}
\captionof{figure}{
\textbf{Training Setup.} Given an input view ($\textbf{I}_\text{1}$, blue) we want to reconstruct the scene, including partially occluded objects such as the green car. To learn to reconstruct both the free space behind the blue car and the surface of the green car, we try to render a pixel of the green view ($\textbf{I}_\text{L}$) by casting a ray and sampling points on it (dotted line). We project the points into the blue view to estimate the density and into the red view ($\textbf{I}_\text{R}$) to sample colors. The pixel of $\textbf{I}_\text{L}$ is only reconstructed correctly if $\textbf{I}_\text{1}$ estimates the correct density of the scene, although the green car and the free space next to it is occluded. In our multi-view method, we propose to use more views, \eg the yellow view $\textbf{I}_\text{2}$ to aggregate more information about the scene to better density predictions. In this example, the yellow view has a slightly better visibility on the green car.
}
\label{fig:training_setup}
\end{figure}

\subsection{Training via Volumetric Rendering}
For convenience, this section recaps the training scheme from \ac{bts} \cite{wimbauer2023behind}, which we follow closely. As with \ac{bts}, this work aims to predict accurate 3D geometry with a continuous scene representation trained only on posed images in a self-supervised manner. We use our geometry prediction and image-based rendering techniques in a differentiable volumetric rendering pipeline to reconstruct images, allowing us to leverage a photometric reconstruction loss to train our networks. See the right-hand side of \autoref{fig:architecture} for an overview of the volumetric rendering. 

\pseudoparagraph{Color Sampling.} Unlike most \ac{nerf} models, we do not predict color that is used to render novel views directly. Instead, we rely on other images to provide us with color samples. Given the point $\textbf{x}$ and a frame $k \in I_{R}$, we obtain a color sample $c_{\textbf{x}, k}$ by projecting $\textbf{x}$ into frame $k$ and interpolating the color value of the image $c_{\textbf{x}, k} = \textbf{I}_k(\textbf{u}^\prime_k)$. 

\pseudoparagraph{Volumetric rendering.} To render a pixel for \ac{nvs} from an image $\textbf{I}_L$, we cast a ray from the camera through the pixel. We sample $M$ discrete points along the ray at positions $\textbf{x}_i$. This discrete sampling lets us approximate the continuous volume rendering equation \cite{mildenhall2021nerf}. For each point, we extract its density $\sigma_{\textbf{x}_i}$ from our multi-view density field and colors $c_{\textbf{x}_i, k}$ from different images $k \in I_R$. We render the color of a pixel by aggregating the sampled colors over the density distribution on the ray
\begin{equation}
    \hat{c}_k = \sum_{i=1}^{M}T_i \alpha_i c_{\textbf{x}_i, k} \,,
\end{equation}
with $\alpha_i$, the probability of the ray ending between $\textbf{x}_i$ and $\textbf{x}_{i + 1}$, $\delta_i$, the distance between $\textbf{x}_i$ and $\textbf{x}_{i + 1}$, and $T_i$, the accumulated transmittance,  given by
\begin{equation}
    \alpha_i = 1 - \exp(- \sigma_{\textbf{x}_i} \delta_i)\quad\quad
    T_i = \prod_{j=1}^{i-1} (1 - \alpha_j)\,.
\end{equation}
Based on the density and $d_i$, the distance between $\textbf{x}_i$ and the ray origin, we can also calculate the expected ray termination depth 
\begin{equation}
    \hat{d} = \sum_{i=1}^{m}T_i \alpha_i d_i\,.
\end{equation}
We want to note that this volumetric rendering reconstructs multiple color suggestions $\hat{c}_k$ for a single pixel based on the different frames $k$, where the color samples come from.

A key insight of \ac{bts} was the setup of the cameras during training. Separating rendering, color sampling, and density prediction to different views allows one to learn the density even in occluded regions of the image. We follow this training setup, adapted to account for multiple input views; see \autoref{fig:training_setup} for more details on this.

\pseudoparagraph{Loss functions.}
We do not reconstruct individual pixels for training the networks, but rather patches of pixels $P_i$, randomly sampled from the loss images $\textbf{I}_L$. Our reconstruction loss is a photometric-consistency loss, combining an L1 and an SSIM loss, following the strategy of \cite{wimbauer2023behind, godard2019digging} of only selecting the per-pixel \textit{minimum} over the reconstructions $\hat{P}_{i, k}$ when aggregating the costs. 
\begin{equation}
    \mathcal{L}_\text{ph} = \min_{k\in I_R} \left(\lambda_\text{L1} \text{L1}(P_i, \hat{P}_{i, k}) + \lambda_\text{SSIM} \text{SSIM}(P_i, \hat{P}_{i, k})\right)
\label{eq:loss_ph}
\end{equation}
Additionally, we also employ an edge-aware smoothness loss term on $d^\star$ an inverse parameterized, mean-normalized version of $\hat{d}$,
\begin{equation}
    \mathcal{L}_\text{eas} = \left|\delta_x d^\star_i\right|e^{-\left|\delta_x P_i\right|} + \left|\delta_y d^\star_i\right|e^{-\left|\delta_y P_i\right|}
\label{eq:loss_eas}
\end{equation}
to regularize the expected ray termination depth, giving us the final loss
\begin{equation}
    \mathcal{L} =  \mathcal{L}_\text{ph} + \lambda_\text{eas}  \mathcal{L}_\text{eas}\,.
\label{eq:loss}
\end{equation}
We give additional details for the training on the different datasets in \autoref{sec:experiments}.
\begin{figure}
\centering
\captionsetup{type=figure}
\includegraphics[trim={0cm 0cm 0cm 0cm},clip,width=\linewidth]{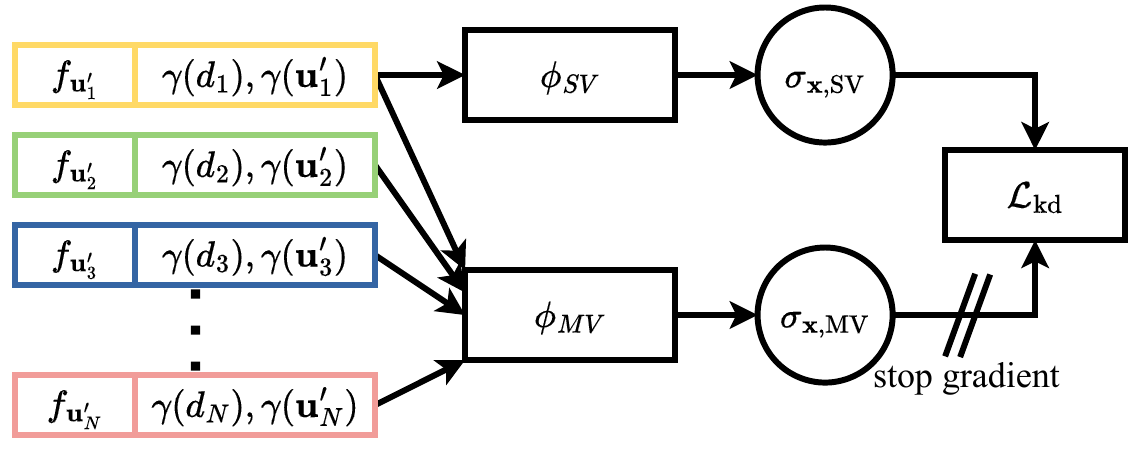}
\vspace{-0.7cm}
\captionof{figure}{
\textbf{Knowledge Distillation.} To improve the single-view (SV) density field reconstruction, we propose leveraging knowledge distillation from the multi-view (MV) predictions. Both $\phi_\text{SV}$ and $\phi_\text{MV}$ make use of the same encoder-decoder architecture and, therefore, the same feature vectors. The knowledge distillation loss $\mathcal{L}_\text{kd}$ pushes the $\phi_\text{SV}$ \ac{mlp} to predict the same density as $\phi_\text{MV}$ while relying only upon a single feature vector. The stop gradient operator prevents $\mathcal{L}_\text{kd}$ influencing $\phi_\text{MV}$.
}
\label{fig:knowledge_distillation}
\vspace{-0.5cm}
\end{figure}

\subsection{KDBTS: Knowledge Distillation for Single-View Reconstruction}
Given the scene reconstructions from our multi-view model, we want to distill this knowledge into a single-view prediction model that can reconstruct the scene from fewer input data, \eg a single image. We use the original \ac{bts} model architecture for our KDBTS. The motivation behind our knowledge distillation approach is threefold. First, it reduces the network size, as the decoder is slightly smaller, speeding up the inference time. Second, removing the necessity of pose information at inference time. The advantage of single-view reconstruction is that it neither requires a calibrated stereo camera setup nor relative pose information from visual odometry systems to reconstruct the scene. Third, it leverages the advantages of self-supervised and supervised training. MVBTS can be trained on image data alone, allowing it to scale to vast amounts of data. This allows our method to be employed in the absence of ground truth data and still provide direct supervision by generating a pseudo ground truth. This knowledge distillation scheme is illustrated in \autoref{fig:knowledge_distillation}.

Given the input images $\textbf{I}_\text{I}$ and $\{\textbf{I}_k\}_{k \in D}$ for single- and multi-view, respectively,  $\sigma_{\textbf{x}, \text{SV}} = \phi_{SV}(f_{\textbf{u}^\prime_\text{I}}, \gamma(d_\text{I}, \textbf{u}^\prime_\text{I}))$ and 
$\sigma_{\textbf{x}, \text{MV}} = \phi_{MV}(\{f_{\textbf{u}^\prime_k}, \gamma(d_k, \textbf{u}^\prime_k)\}_{k \in I_D})$ 
should give the same density prediction for a 3D point $\textbf{x}$. In practice, we choose the input view $\textbf{I}_\text{I}$ for the single-view head to be from $\{\textbf{I}_k\}_{k \in D}$, but this is not strictly necessary. Knowledge distillation is then performed via a simple L1 loss
\begin{equation}
    \mathcal{L}_{\text{kd}} = \|\sigma_{\textbf{x}, \text{MV}} - \sigma_{\textbf{x}, \text{SV}}\|\,.
\end{equation}
To avoid the multi-view head being influenced by the single-view prediction, we apply a stop gradient operator on $\sigma_{\textbf{x}, \text{MV}}$ to treat it as a pseudo ground truth.

\begin{table}
\footnotesize
	\centering
\begin{tabular}{l c lll}
\toprule
\scriptsize\textit{Dataset} & \scriptsize\hspace{-.2cm} Split.\ \hspace{-.2cm} & \scriptsize \# Train & \scriptsize \# Val & \scriptsize\hspace{-.2cm} \# Test \hspace{-.2cm} \\
\midrule
KITTI        & Eigen ~\cite{eigen2014depth}   & 39810             & 4424            & 697              \\
\midrule
KITTI-360        & BTS ~\cite{wimbauer2023behind} & 98008             & 11451           & 446              \\
\bottomrule
\end{tabular}
\caption{\textbf{Dataset Split Overview.} Number of images in each split of the datasets. Evaluation is performed exclusively on the test splits.}
\vspace{-.5cm}
\label{tab:dataset_overview}
\end{table}

\section{Experiments}
\label{sec:experiments}

\begin{figure*}
\centering
\captionsetup{type=figure}
\includegraphics[trim={0cm 0cm 0cm 0cm},clip,width=\linewidth]{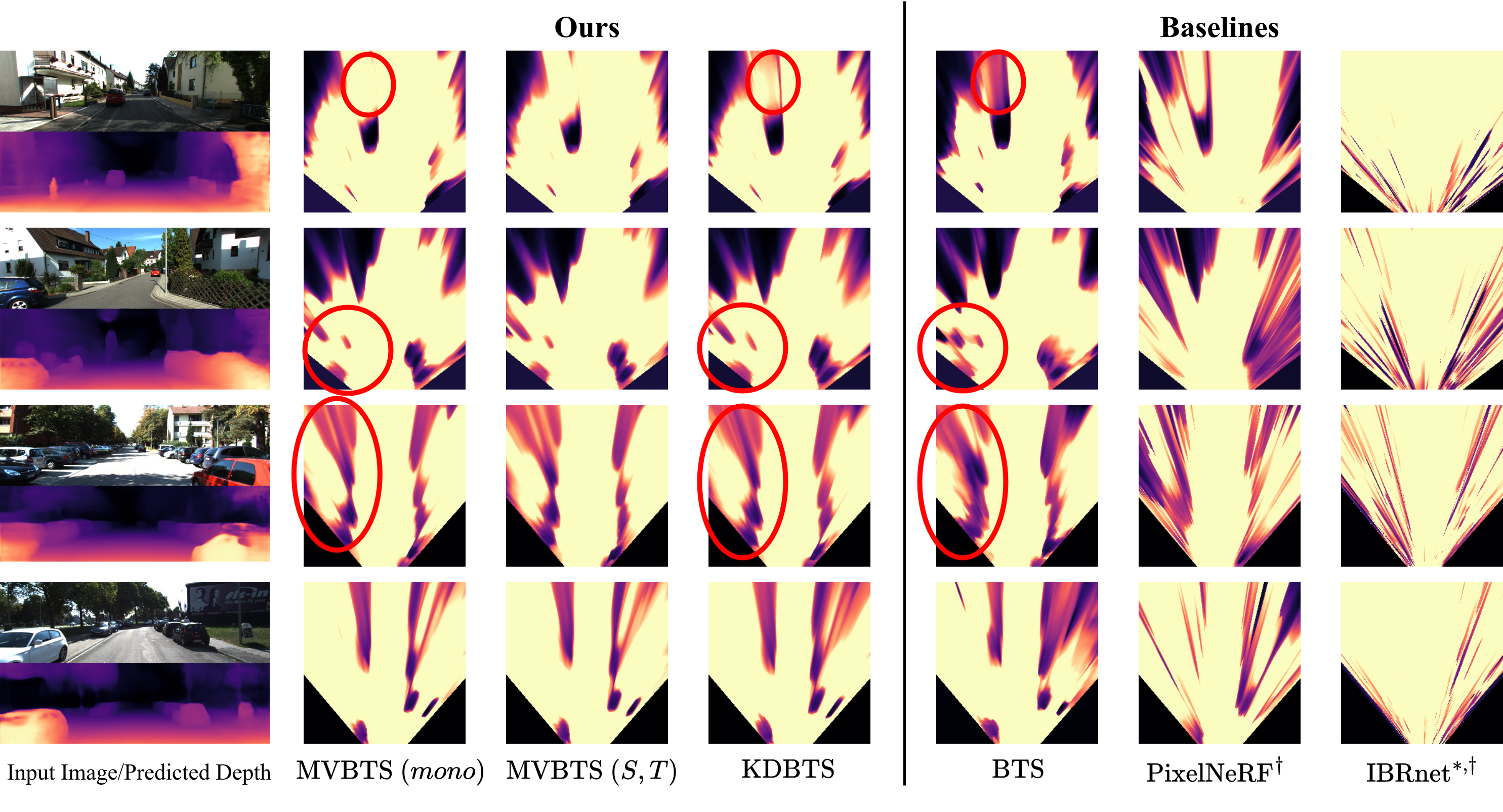}
\vspace{-0.8cm}
\captionof{figure}{\textbf{Density Fields.} Top-down rendering of the density fields for an area of $x = \left[-9m,9m\right]$, $y = \left[0m,1m\right]$, $z = \left[3m,23m\right]$. Images are taken from KITTI-360 (top half) and KITTI (bottom half) with profiles coming from models trained on KITTI-360. Every model except for MVBTS $(S, T)$ and IBRnet \cite{wang2021ibrnet} get the same input data. Our MVBTS can predict accurate geometry even in distant regions for both a single image and multiple images. KDBTS learns to recreate the accurate density structure from MVBTS. Both models reduce the amount of shadows produced by \ac{bts} \cite{wimbauer2023behind}, especially in distant regions. They also produce cleaner boundaries for close-by objects. Note that KDBTS has a smaller model capacity than MVBTS $(mono)$. $*$: changed sensitivity for visualization purposes, $\dagger$: retrained on KITTI-360. 
}
\label{fig:profiles}
\vspace{-0.4cm}
\end{figure*}

We demonstrate the advantages of having multiple views to predict density fields in the tasks of \textit{depth prediction} and \textit{occupancy estimation}. We evaluate both our multi-view model (MVBTS) as well as our single-view model (KDBTS) boosted by knowledge distillation. We follow the evaluation of \cite{wimbauer2023behind} and evaluate the depth prediction on the KITTI \cite{geiger2013vision} dataset against the ground truth depth and occupancy estimation on KITTI-360 \cite{liao2022kitti} against occupancy maps from aggregated LiDAR scans over multiple time steps. 

\subsection{Datasets and Training Details}

KITTI \cite{geiger2013vision} and KITTI-360 \cite{liao2022kitti} are both autonomous driving datasets that provide time-stamped stereo images with ground truth poses - although for KITTI we use poses generated by ORB-SLAM 3 \cite{campos2021orb}. KITTI-360 also provides fisheye cameras looking to the sides. For our training, they are rectified to the same camera intrinsics as the stereo cameras. A randomly applied dropout of $0.5$, helps our model to not overfit on a specific camera setup. During training, we randomly select three stereo input views to be in $I_D$. 

We train our model on a single 48GB Nvidia RTX A40 GPU with half the batch size of BTS, 8, due to increased memory demand from the multiple views. We sample 32 patches of size $8\times 8$ with 64 points per ray for possible reconstruction. The encoder-decoder follows \cite{wimbauer2023behind} to be a ResNet50 encoder \cite{he2016deep} with 64 output channels. $\text{MLP}_1$ and $\text{MLP}_2$ are chosen to be two-layer networks with residual connections. $\text{MLP}_1$ therefore corresponds to a larger version of the decoder network of \ac{bts}, as it needs to decode additional information. For detailed hyperparameters, please refer to the supplementary material.

We train for 150,000 steps on KITTI and for 200,000 steps on KITTI-360, compensating for our smaller batch size. We train on the splits detailed in \autoref{tab:dataset_overview}.

In our experimental setup, both KDBTS and MVBTS share a backbone architecture. To train the KDBTS model, we use a trained MVBTS model and freeze the encoder-decoder backbone as well as $\phi_{MV}$. We found training both in parallel to give a slightly decreased performance. To additionally boost the performance in distant parts of the scene, we randomly sample the timestamp offset of the additional stereo frames between $0.1s$ and $0.8s$. As direct supervision and a shared backbone lead to a strong training signal, we train the knowledge distillation for only 20,000 steps.

\subsection{Depth Prediction}
\begin{figure}
\centering
\captionsetup{type=figure}
\includegraphics[trim={0cm 0cm 0cm 0cm},clip,width=\linewidth]{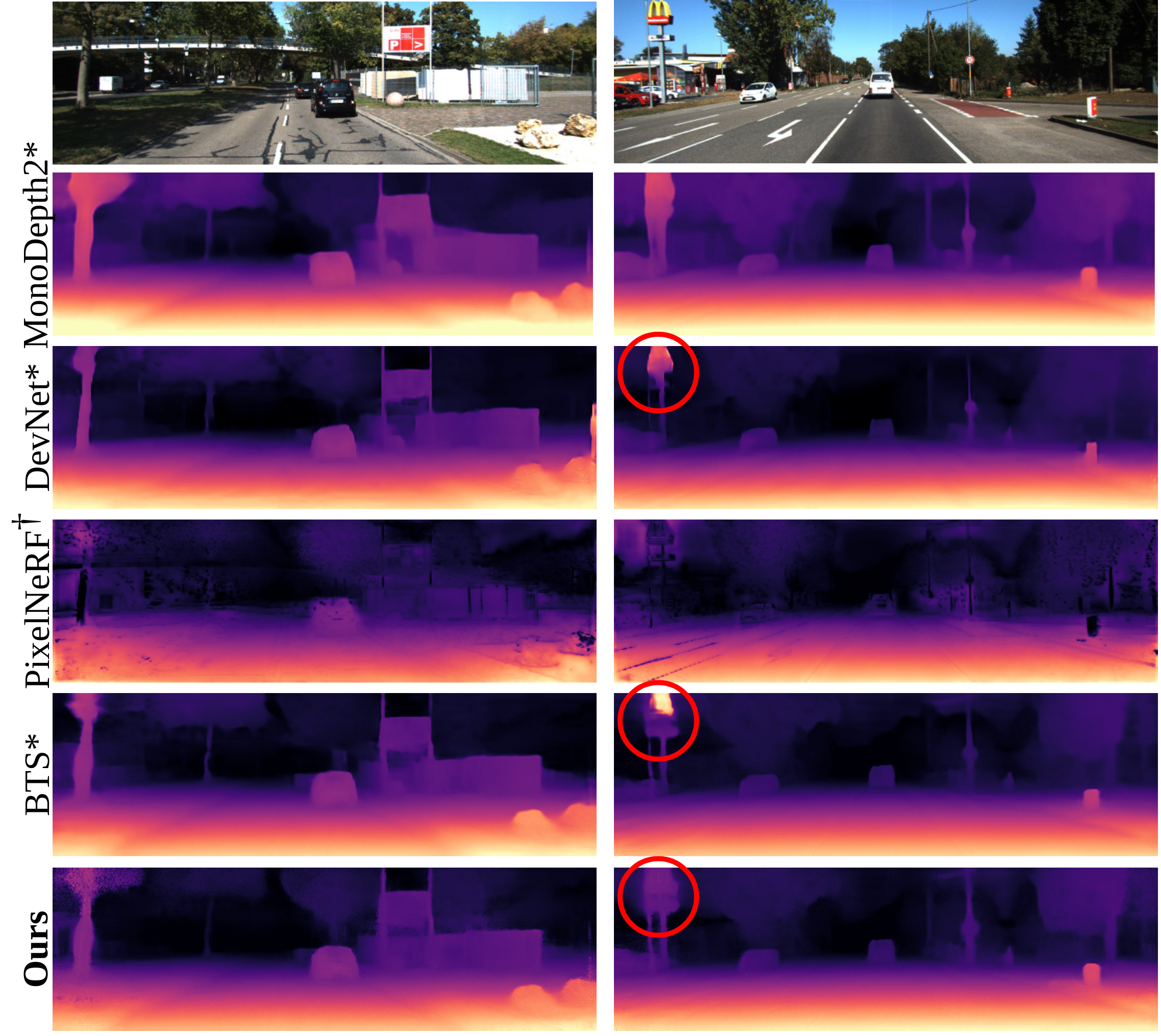}
\captionof{figure}{
\textbf{Monocular Depth Prediction.} Qualitative comparison with state-of-the-art monocular depth prediction and other volumetric methods. The expected ray termination depth $\hat{d}$ gives a detailed scene reconstruction. 
}
\label{fig:depth_prediction}
\vspace{-0.5cm}
\end{figure}

Although a byproduct of the density field, depth estimates provide useful information leveraged by downstream tasks, such as segmentation. In \autoref{tab:depth_pred}, we compare our two methods against self-supervised depth prediction methods and volume reconstruction methods. \autoref{fig:depth_prediction} shows qualitative results on KITTI. For depth prediction, we keep the same training setup for KITTI so that we report the values for PixelNeRF \cite{yu2021pixelnerf}, MonoDepth2 \cite{godard2019digging}, DevNet \cite{zhou2022devnet}, and \ac{bts} \cite{wimbauer2023behind} from \cite{wimbauer2023behind}. For more single-view methods, please refer to \cite{wimbauer2023behind}.

\begin{table}
\centering
\footnotesize
\begin{tabular}{lcccc}
\toprule
\scriptsize\textit{Model} & \scriptsize\hspace{-.2cm}Volum.\ \hspace{-.2cm} & \scriptsize Abs Rel $\downarrow$ & \scriptsize RMSE $\downarrow$ & \scriptsize\hspace{-.2cm}$\alpha < 1.25 \uparrow$\hspace{-.2cm} \\
\midrule
PixelNeRF \cite{yu2021pixelnerf} & \cmark & 0.130 & 5.134 & 0.845 \\
MonoDepth2 \cite{godard2019digging}\hspace{-.2cm} & \xmark & 0.106 & 4.750 & 0.874 \\
DevNet  \cite{zhou2022devnet}    & (\cmark) & \textbf{0.095} & \textbf{4.365} & \textbf{0.895} \\
BTS \cite{wimbauer2023behind} & \cmark  & \underline{0.102} & \underline{4.407} & \underline{0.882} \\ 
\midrule
KDBTS (\textbf{Ours}) & \cmark & 0.105 & 4.497 & 0.873 \\
MVBTS (\textbf{Ours}) & \cmark & 0.105 & 4.501 & 0.873 \\
\bottomrule
\end{tabular}
\caption{\textbf{Monocular Depth Evaluation.} Our method performs very similar to \ac{bts} due to the similarity of the architecture. We suspect that the slight differences in the presented metrics arise from some of the model capacity being shifted to better predict the occluded parts of the scene.}
\vspace{-.2cm}
\label{tab:depth_pred}
\end{table}

\begin{table*}[t]
\centering
\footnotesize
\begin{tabular}{lc|ccc|ccc}
\toprule
\textit{Method} & Multi View & $\text{O}_\text{acc} \uparrow$ & $\text{O}_\text{prec} \uparrow$ & $\text{O}_\text{rec} \uparrow$ & $\text{IE}_\text{acc} \uparrow$  & $\text{IE}_\text{prec}$ & $\text{IE}_\text{rec} \uparrow$ \\
\midrule
PixelNeRF \cite{yu2021pixelnerf}  & \xmark & 93.82\% & 	51.94\% & 	69.43\% & 	61.33\% & 	37.86\% & 	42.21\%  \\
BTS \cite{wimbauer2023behind} & \xmark & 94.47\% & 	58.73\% & 	84.24\% & 	77.04\% & 	\textbf{54.21\%} & 	43.99\%   \\
\midrule
MVBTS (\textbf{Ours}) & \xmark & \textbf{94.76\%} & 	\textbf{60.83\%} & 	\underline{84.51\%} & 	\underline{78.00\%} & 	\underline{53.69\%} & 	\textbf{44.04\%}  \\
KDBTS (\textbf{Ours}) & \xmark & \textbf{94.76\%} & 	\underline{60.68\%} & 	\textbf{84.78\%} & 	\textbf{78.30\%} & 	53.62\% & 	\underline{44.00\%}  \\
\midrule
\midrule
IBRnet \cite{wang2021ibrnet}  & \cmark & \underline{96.03\%} & 	4.14\% & 	4.78\% & 	34.36\% & 	32.97\% & 	\textbf{96.02\%}  \\
IBRnet (depth + 4m) \cite{wang2021ibrnet}  & \cmark & \textbf{98.12\%} & 	\underline{43.35\%} & 	\textbf{86.01\%} & 	\underline{59.67\%} & 	\underline{25.18\%} & 	9.85\%  \\
\midrule
MVBTS (\textbf{Ours}) & \cmark & 94.91\% & 	\textbf{61.73\%} & 	85.78\% & 	\textbf{79.47\%} & 	\textbf{55.08\%} & 	\underline{45.23\%}  \\
\bottomrule
\end{tabular}
\vspace{-.2cm}
\caption{\textbf{Occupancy Prediction on KITTI-360.} We follow the evaluation setup of BTS \cite{wimbauer2023behind} and evaluate both the occupancy of the whole scene ($\text{O}$) and only the invisible and empty parts ($\text{IE}$). In both cases, we report accuracy, precision, and recall, as there is a large imbalance between occupied and empty parts of the scene. We evaluate MVBTS and KDBTS in a single-view and MVBTS additionally in a multi-view setting. KDBTS can improve upon the original BTS consistently. Due to IBRnet \cite{wang2021ibrnet} predicting little density (see \autoref{fig:profiles}), we also evaluate IBRnet in the setting of depth + 4m, giving better occupancy precision and recall.}
\vspace{-.2cm}
\label{tab:occ_pred}
\end{table*}

\begin{table*}[t]
\centering
\footnotesize
\begin{tabular}{lcccc|ccc|ccc}
\toprule
Inference & dropout & $\text{MLPs}$ & attn. layers & encode fisheye & $\text{O}_\text{acc} \uparrow$ & $\text{O}_\text{prec} \uparrow$ & $\text{O}_\text{rec} \uparrow$ & $\text{IE}_\text{acc} \uparrow$ & $\text{IE}_\text{prec}$ & $\text{IE}_\text{rec} \uparrow$ \\
\midrule
(\textit{S}, \textit{T}) & 0.5 & middle & \textcolor{Green}{\cmark} & \textcolor{Red}{\xmark} & \textcolor{Red}{-1.67\%} & \textcolor{Red}{-0.96\%} & \textcolor{Red}{-15.09\%} & \textcolor{Red}{-9.64\%} & \textcolor{Red}{-7.67\%} & \textcolor{Green}{13.70\%} \\
(\textit{S}, \textit{T}) & 0.0 & middle & \textcolor{Red}{\xmark} & \textcolor{Green}{\cmark} & \textcolor{Red}{-1.47\%} & \textcolor{Red}{-4.88\%} & \textcolor{Red}{-6.04\%} & \textcolor{Red}{-6.45\%} & \textcolor{Red}{-9.54\%} & \textcolor{Red}{-1.33\%} \\
(\textit{S}, \textit{T}) & 0.2 & middle & \textcolor{Red}{\xmark} & \textcolor{Green}{\cmark} & \textcolor{Red}{-0.34\%} & \textcolor{Red}{-1.88\%} & \textcolor{Red}{-2.58\%} & \textcolor{Red}{-2.34\%} & \textcolor{Red}{-0.67\%} & \textcolor{Green}{0.41\%} \\
(\textit{S}, \textit{T}) & 0.5 & middle & \textcolor{Red}{\xmark} & \textcolor{Green}{\cmark} & \textcolor{Green}{0.02\%} & \textcolor{Red}{-1.01\%} & \textcolor{Red}{-0.31\%} & \textcolor{Red}{-0.42\%} & \textcolor{Green}{0.65\%} & \textcolor{Green}{1.16\%} \\
(\textit{S}, \textit{T}) & 0.8 & middle & \textcolor{Red}{\xmark} & \textcolor{Green}{\cmark} & \textcolor{Green}{0.04\%} & \textcolor{Red}{-1.30\%} & \textcolor{Green}{0.62\%} & \textcolor{Red}{-0.06\%} & \textcolor{Green}{0.27\%} & \textcolor{Red}{-0.43\%} \\
(\textit{S}, \textit{T}) & 0.5 & small & \textcolor{Red}{\xmark} & \textcolor{Red}{\xmark} & \textcolor{Red}{-0.26\%} &  \textcolor{Red}{-2.09\%} &  \textcolor{Green}{0.86\%} &  \textcolor{Red}{-0.93\%} &  \textcolor{Red}{-0.04\%} &  \textcolor{Red}{-1.53\%} \\
(\textit{S}, \textit{T}) & 0.5 & large & \textcolor{Red}{\xmark} & \textcolor{Red}{\xmark} & \textcolor{Red}{-0.53\%} &  \textcolor{Red}{-1.84\%}&  \textcolor{Red}{-3.27\%}&  \textcolor{Red}{-2.23\%}&  \textcolor{Red}{-3.41\%} &  \textcolor{Green}{1.66\%} \\
\midrule
\midrule
(\textit{Mono}) & 0.5 & middle & \textcolor{Red}{\xmark} & \textcolor{Red}{\xmark} & \textcolor{Red}{-0.15\%} & \textcolor{Red}{-0.90\%} & \textcolor{Red}{-1.27\%} & \textcolor{Red}{-1.47\%} & \textcolor{Red}{-1.38\%} & \textcolor{Red}{-1.19\%} \\
(\textit{S}) & 0.5 & middle & \textcolor{Red}{\xmark} & \textcolor{Red}{\xmark} & \textcolor{Red}{-0.07\%} & \textcolor{Red}{-0.61\%} & \textcolor{Red}{-0.50\%} & \textcolor{Red}{-0.85\%} & \textcolor{Red}{-1.06\%} & \textcolor{Red}{-1.19\%} \\
(\textit{T}) & 0.5 & middle & \textcolor{Red}{\xmark} & \textcolor{Red}{\xmark} & \textcolor{Red}{-0.09\%} & \textcolor{Red}{-0.77\%} & \textcolor{Red}{-0.88\%} & \textcolor{Red}{-0.63\%} & \textcolor{Red}{-1.01\%} & \textcolor{Red}{-0.04\%} \\
(\textit{S}, \textit{T})& 0.5 & middle & \textcolor{Red}{\xmark} & \textcolor{Red}{\xmark} & 94.91\% & 61.73\% & 85.78\% & 79.47\% & 55.08\% & 45.23\% \\
\bottomrule
\end{tabular}
\vspace{-0.2cm}
\caption{\textbf{Ablation Studies.} To motivate the choice of our architecture for \ac{mvbts}, we detail the results for different settings. We also report the influence of additional images at inference time. Overall, the architecture is robust against the most hyperparameter settings. The ablation study shows that keeping the multi-view aggregation simple. We specifically address the use of attention layers together with a detailed breakdown of the different settings in the supplementary material.}
\vspace{-.6cm}
\label{tab:occ_ablation}
\end{table*}

\subsection{Occupancy Prediction}
The main evaluation of our method is with regard to the occupancy prediction. Next to our method, we evaluate with PixelNeRF \cite{yu2021pixelnerf} and the original \ac{bts} \cite{wimbauer2023behind} two additional single-view reconstruction methods. For multi-view methods, we compare against another image-based rendering method, IBRnet \cite{wang2021ibrnet} retrained on KITTI-360. \autoref{fig:profiles} shows top-down renderings of occupancy predictions both on KITTI and KITTI-360 made with models trained on KITTI-360. They show qualitative results for difficult scenarios. For IBRnet, we increased the sensitivity of the visualization as the density values were, on average, significantly lower than for the other methods. Our models reason better in far-away parts of the scene (see first example). Additionally, they produce cleaner occupancy edges in occluded areas of the scene, leading to fewer artifacts. For quantitative evaluation, we follow \cite{wimbauer2023behind} with carving out free space based on LiDAR scans. We evaluate accuracy, precision, and recall for two settings: 1.\ Occupancy predictions ($\text{O}_\text{acc}$, $\text{O}_\text{prec}$, $\text{O}_\text{rec}$) and 2.\ Invisible and empty ($\text{IE}_\text{acc}$, $\text{IE}_\text{prec}$, $\text{IE}_\text{rec}$), giving us six metrics in total. \autoref{tab:occ_pred} details the results on KITTI-360. Unlike \ac{bts} \cite{wimbauer2023behind}, we also choose to include the precision and recall for the occupancy due to the large imbalance between occupied and empty space. As the distribution is heavily skewed to the empty space, achieving a high accuracy score is easy, as can be seen for IBRnet. On KITTI-360, IBRnet seems to give fog-like predictions with few clear boundaries, as can be seen in \autoref{fig:profiles}, resulting in low precision and recall. We also evaluate IBRnet in the depth + $4m$ setting to address this. This reduces the $\text{IE}_\text{rec}$ but gives competitive other metrics. Our models, both \ac{mvbts} and KDBTS, are evaluated in the single-view setting in the top half of \autoref{tab:occ_pred}. Both methods consistently improve the original \ac{bts} across most metrics. Despite the lower model capacity, KDBTS performs on par with MVBTS, validating our knowledge distillation approach. In the multi-view setting, where the models have access to four frames - stereo and temporal -, MVBTS performs the most consistently across all metrics. IBRnet has a better $\text{O}_\text{acc}$ and, depending on the prediction mode, either a slightly better $\text{O}_\text{rec}$ or significantly better $\text{IE}_\text{rec}$. We suspect this results from the large imbalance of empty to occupied space. MVBTS either performs on par or significantly better than IBRnet. We present additional evaluation results in the supplementary material.

\subsection{Ablation Studies}
\autoref{tab:occ_ablation} gives an overview of the impact different design choices in our architecture and training scheme have on the performance of our model. Our ablation study is split into two parts. In the first part, we are concerned with the architecture of $\phi_{MV}$, whereas ablations on the backbone have already been done in \cite{wimbauer2023behind}. The second part investigates the influence of the number of available views on the performance of the model. For our ablation study, we use our final model as a baseline and report the differences to this model in the table. We investigate the following: adding attention layers similar to GeoNeRF \cite{johari2022geonerf} for enhanced information sharing between views; also including fisheye views in $I_D$; the effects of different dropout rates; the size of the \acp{mlp}. In general, our method shows robust performance in the ablation settings. Attention layers allow for information sharing between tokens and can implicitly learn to shift focus to certain views. However, within this architecture, they seem to hinder the model's performance. We detail a possible explanation together with examples in the supplementary material. Including fisheye cameras leads to a mostly on-par performance. This is likely due to most sampled points lying outside of the fisheye cameras' frustums. Less dropout leads to worse generalization capabilities in the network, which can especially be seen in the monocular setting reported in the supplementary material. Directly summing densities from different images ($\text{MLPs}$: small) also leads to a decreased performance. An evaluation of the number of images at inference time concludes the ablation study. Including both stereo and temporal frames seems to give a similar boost in performance, whereas having both available results in the most accurate predictions.

\section{Discussion}
\label{sec:discussion}

Our method for accurately reconstructing density fields relies on several assumptions. In the following, we will discuss these assumptions and their possible effects on the reconstruction quality. As with many \ac{nerf} or scene reconstruction methods - with the exception of dynamic \acp{nerf}, \etc -, our method uses the static scene assumption. However, datasets such as KITTI contain moving objects in the form of cars, pedestrians, and cyclists. Dynamic objects influence both the training of \ac{mvbts} and KDBTS during knowledge distillation, possibly resulting in drawn-out shadows of objects. We share this limitation with other methods that require images from multiple time steps during training and inference. Similar to the static scene assumption, sampling color assumes photo consistency for reconstruction that does not model view-dependent effects of different materials. Methods such as IBRnet and Neuray\cite{liu2022neuray} are able to model these effects, resulting in better novel view synthesis. However, we argue that this leads to a worse geometric reconstruction as shown in \autoref{tab:occ_pred} with PixelNeRF and IBRnet. More images during inference time give \ac{mvbts} more computational overhead compared to BTS. Additionally, our multi-view model needs - at least - relative camera poses at inference time as well. Our single-view model can address these last limitations while simultaneously giving better occupancy predictions than comparable models such as \ac{bts}.

\section{Conclusion}
\label{sec:conclusion}
We introduce a novel approach to enhancing single-view geometric scene reconstruction by leveraging multi-view information. This involves extending a state-of-the-art density prediction model to improve scene geometry, followed by direct supervision in 3D via knowledge distillation to boost a single-view model. Training is done fully self-supervised on video data. We evaluate both the proposed multi-view and boosted single-view models on depth estimation and occupancy prediction tasks. While our method performs close to state-of-the-art for depth estimation, being outperformed by methods explicitly trained for this task, our boosted single-view reconstruction model consistently achieves state-of-the-art performance for occupancy prediction. Future work on modeling moving objects can address conflicting information in dynamic scenes, improving overall accuracy and reliability in 3D reconstructions.

\paragraph{Acknowledgements.}
This work was supported by the ERC Advanced Grant SIMULACRON, by the Munich Center for Machine Learning, and  by the German Federal Ministry of Transport and Digital Infrastructure (BMDV) under grant 19F2251F for the ADAM project.

\newpage
{
    \small
    \bibliographystyle{ieeenat_fullname}
    \bibliography{main}
}

\clearpage
\setcounter{page}{1}
\maketitlesupplementary

\section{Additional Results}
In the following, we present additional results of our methods and further comparisons to existing methods, such as PixelNeRF \cite{yu2021pixelnerf} and \ac{bts}. \autoref{sup:attn_layers} discusses the results of the attention-based model of the ablation study more extensively. \autoref{sup:occ_est} gives more results for our ablation study, detailing the influence of the different inference setups on the performance of the multi-view model. \autoref{sup:occ_profiles} presents more qualitative examples of the occupancy profiles to show both the benefits of our training setup and the influence of additional frames. \autoref{sup:depth_pred} gives additional qualitative results for the depth prediction task on KITTI \cite{geiger2013vision}.

\subsection{Using attention layers}\label{sup:attn_layers}
\begin{figure}
\centering
\captionsetup{type=figure}
\includegraphics[trim={0cm 0cm 0cm 0cm},clip,width=0.13\textwidth]{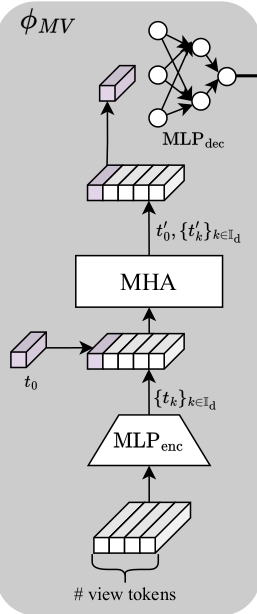}
\captionof{figure}{
\textbf{Attention Layer.} We additionally investigated Multi-Head-Attention (MHA) layers to fuse geometry information from multiple images. As another method of aggregating multi-views, the Multi-Head self-At, replacing the $\Phi_{MV}$ in \autoref{fig:architecture}. \autoref{fig:attn_infer} shows qualitative results with using the attention layers for multi-view aggregation.
}
\label{fig:attn_layer}
\end{figure}

\begin{figure*}
\centering
\captionsetup{type=figure}
\includegraphics[trim={0cm 0cm 0cm 0cm},clip,width=\textwidth]{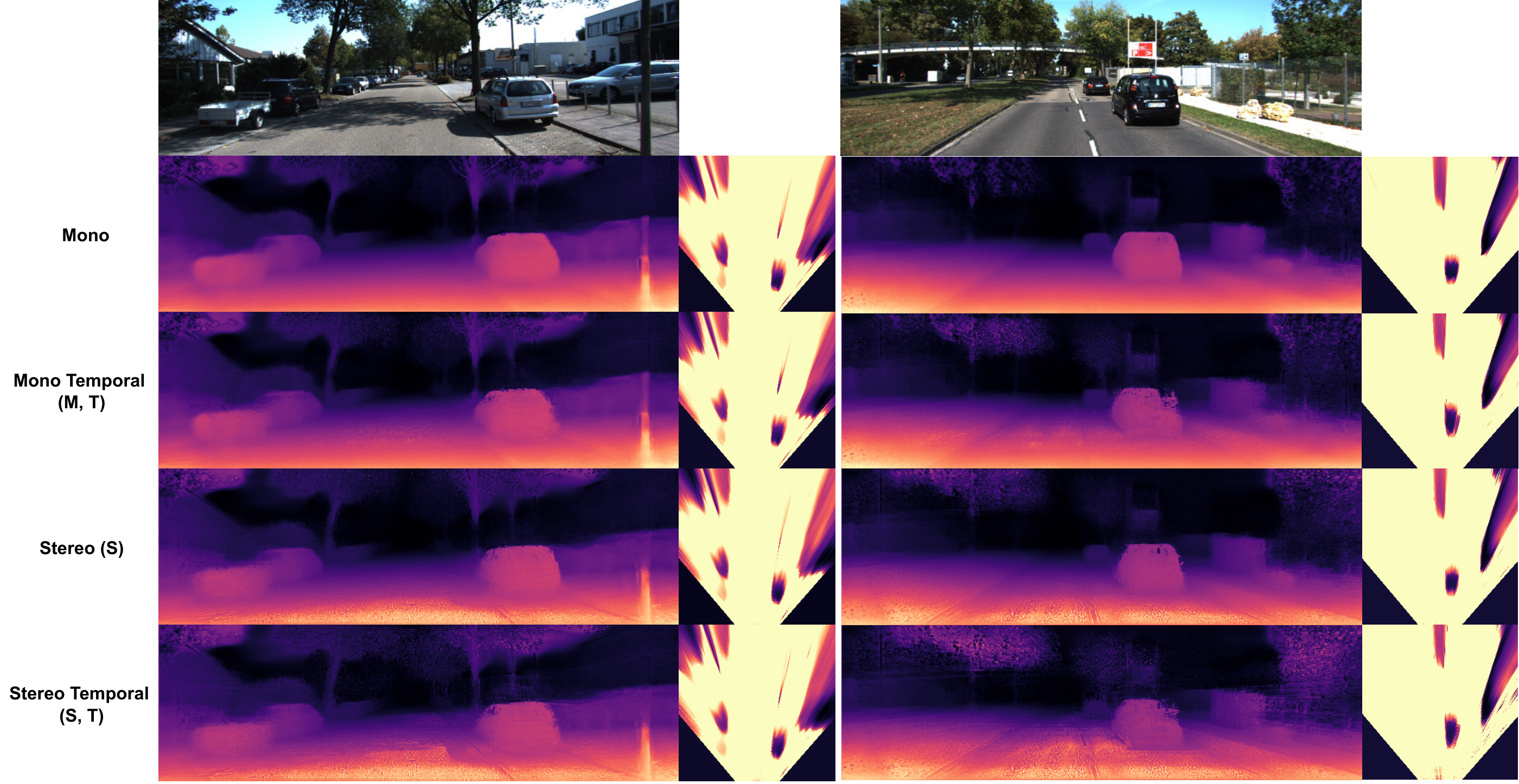}
\captionof{figure}{
\textbf{Attention Layer Qualitative Results.}
		Comparison in both depth- and occupancy estimation. The camera frustum is set up in x=[-9, 9]m, y=[0, 0.75]m, and z=[3,21]m. The monocular occupancy prediction produces reasonable results for both the expected ray termination depth and the occupancy profiles. Adding more frames to the prediction leads to increased noise in the predictions. A closer inspection of the occupancy profiles shows that the attention model produces long and thin \textit{occupancy shadows} along rays cast from the camera. The occupancy predictions seem to be quite sensitive to changes in the features coming from the pixel-aligned feature map.
}
\label{fig:attn_infer}
\end{figure*}

In \autoref{fig:attn_infer}, we show prediction examples coming from the attention model instead of softmax view-aggregation in \autoref{fig:architecture}. (See \autoref{fig:attn_layer}) While the single model seems to produce reasonable depth and occupancy prediction, adding more views leads to noisy depth predictions that get worse with each additional view. A closer inspection of the occupancy profiles shows that for all inference setups, the attention model casts thin \textit{occupancy shadows} in the scene, which seems to degrade the quality of the depth prediction and the occupancy evaluation.
\newpage

\begin{figure}
\centering
\captionsetup{type=figure}
\includegraphics[trim={0cm 0cm 0cm 0cm},clip,width=\linewidth]{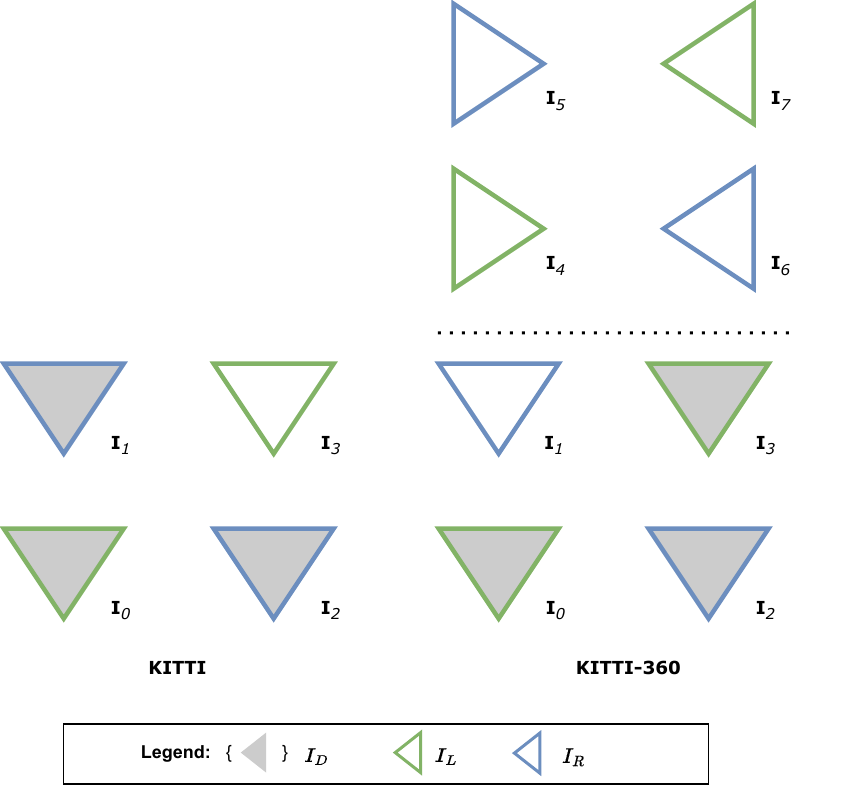}
\captionof{figure}{
\textbf{Frame Arrangement in temporal steps.}
		The possible frames are split into either loss set $I_{L}$ or render loss set $I_{R}$. Note that the first frame starting with zero index is used for an input frame as fixed.
		Both KITTI-360 and KITTI use stereo cameras. Depending on the experiments' setup, having fisheyes as input scenes is optional. This influence has been discussed in the ablation study of the main paper.
}
\label{fig:frame_arrangement}
\end{figure}

\subsection{Occupancy Estimation}\label{sup:occ_est}
\begin{figure}
\centering
\captionsetup{type=figure}
\includegraphics[trim={0cm 0cm 0cm 0cm},clip,width=\linewidth]{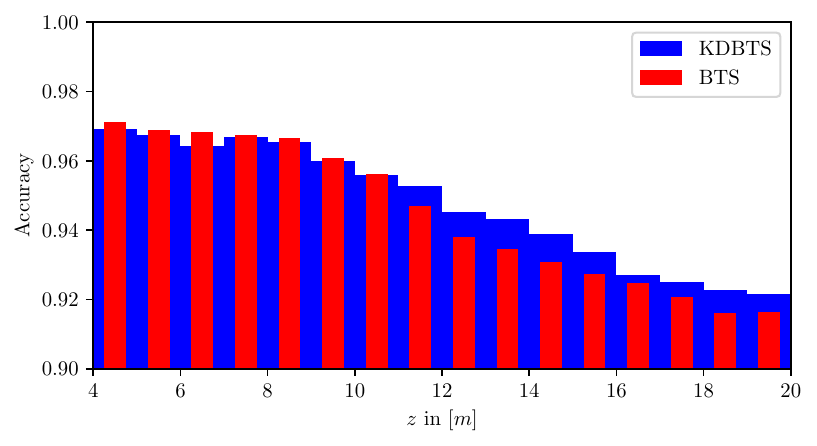}
\vspace{-0.5cm}
\caption{
\textbf{Occupancy Accuracy At Different Depths.} \textcolor{blue}{\acs{kdbts}} shows on par performance for occupancy estimation to \textcolor{red}{\acs{bts}} for depth values (z) smaller than $10$ m. For depths larger than $10$ m, KDBTS starts to outperform BTS significantly.
}
\label{fig:occ_acc_z_values}
\vspace{-0.5cm}
\end{figure}

\autoref{fig:occ_acc_z_values} shows the occupancy accuracy for \ac{kdbts} and \ac{bts} for different depth values. The performance increase of \ac{kdbts} mainly happens at depth values larger than $10$ meters.

\begin{figure*}
\centering
\captionsetup{type=figure}
\includegraphics[trim={0cm 0cm 0cm 0cm},clip,width=0.8\linewidth]{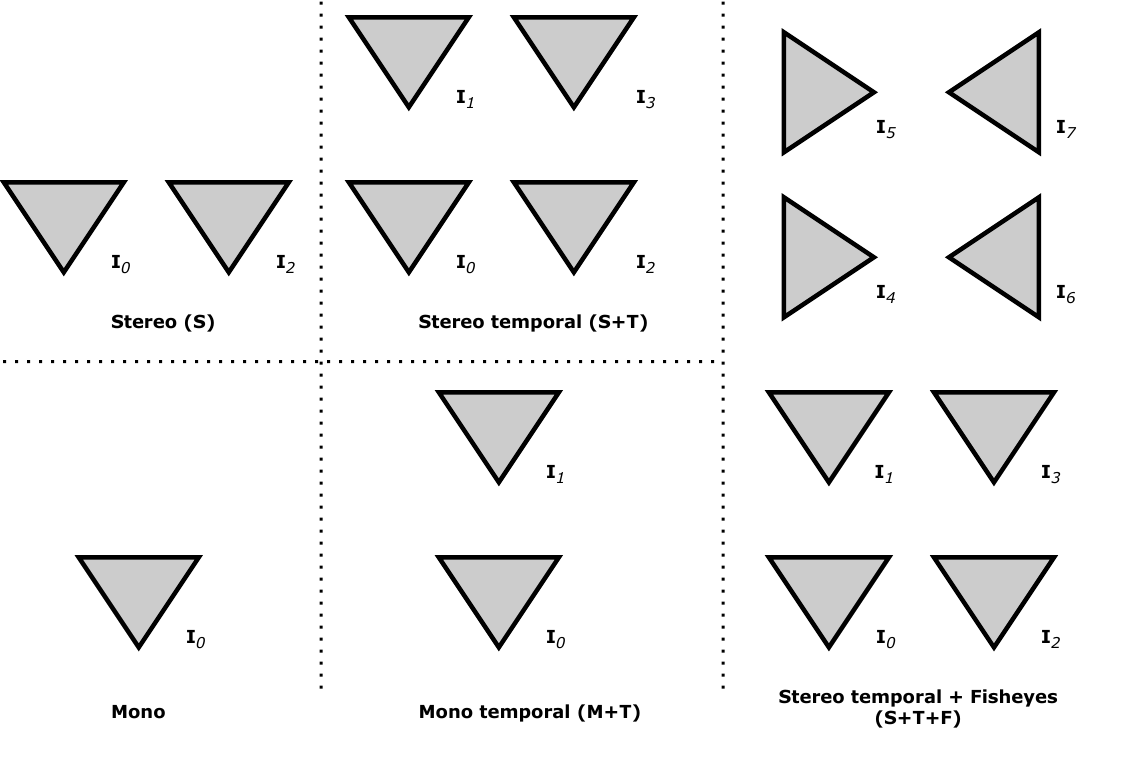}
\captionof{figure}{
\textbf{Frame Arrangement in Inference.} Illustration of the settings used at inference time in \autoref{tab:sup_occ_ablation_mono}, \autoref{tab:sup_occ_ablation_mono_temporal}, \autoref{tab:sup_occ_ablation_stereo}, \autoref{tab:sup_occ_ablation_stereo_temporal}, and \autoref{tab:sup_occ_ablation_stereo_temporal_fisheye}.
}
\label{fig:infer_frame_arrangement}
\end{figure*}

\autoref{tab:sup_occ_ablation_mono} to \autoref{tab:sup_occ_ablation_stereo_temporal_fisheye} gives the complete overview of the settings tested in the ablation study of the main paper for five different camera settings at inference time. The different setups used in these tables are illustrated in \autoref{fig:infer_frame_arrangement}. It shows the general improvement of all models tested in the ablation study when providing more frames at inference time, except for the model with attention layers. Our model performs best in all settings with a few minor exceptions. We additionally show the influence of including fisheye cameras at inference time in \autoref{tab:sup_occ_ablation_stereo_temporal_fisheye}.

\subsection{Occupancy profiles}\label{sup:occ_profiles}
\begin{figure*}
\centering
\captionsetup{type=figure}
\includegraphics[trim={0cm 0cm 0cm 0cm},clip,width=\textwidth]{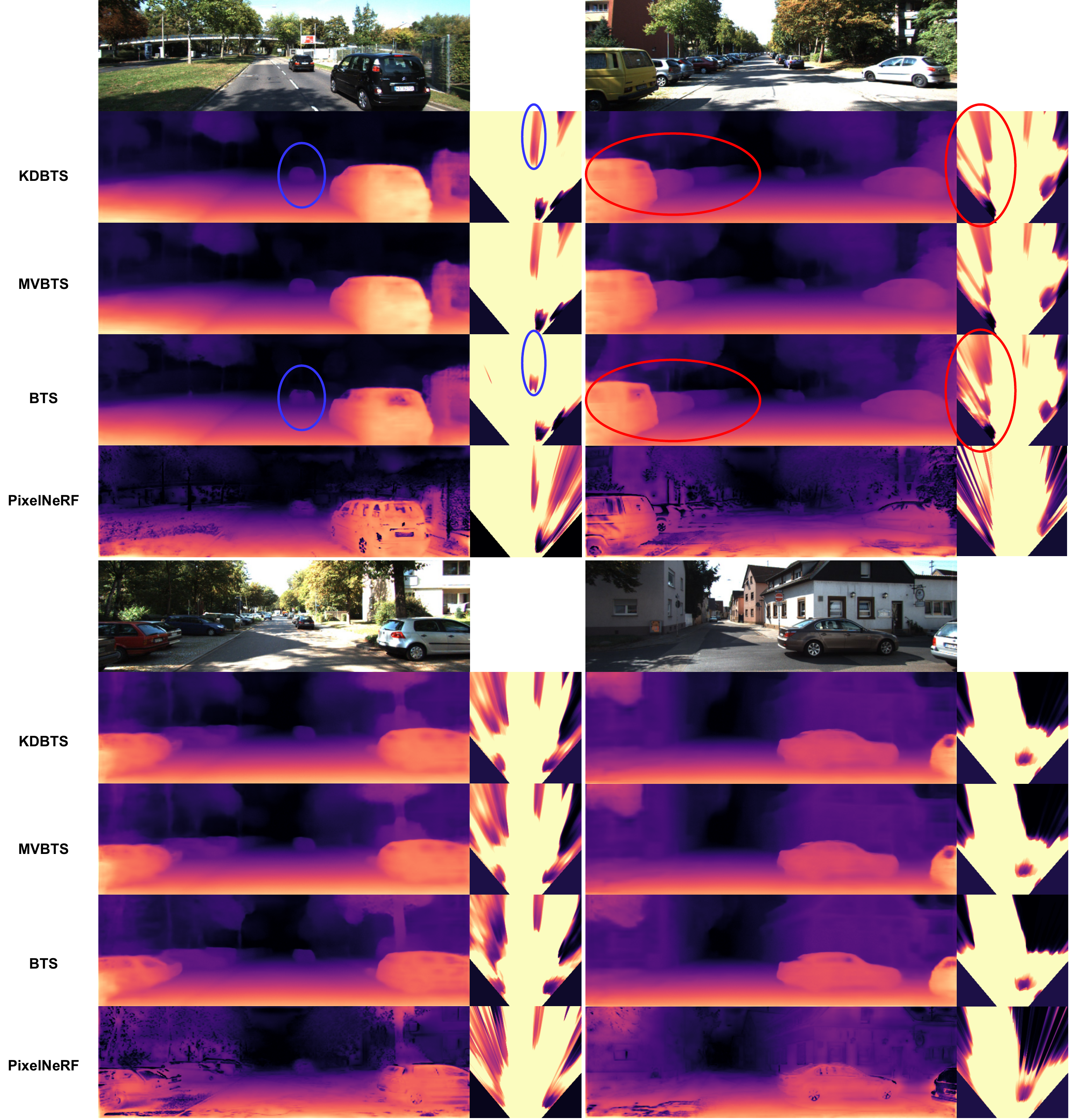}
\captionof{figure}{
\textbf{Qualitative Baseline Results 1.}
		Baselines comparison in both depth- and occupancy estimation. The camera frustum is set up in x=[-9, 9]m, y=[0, 0.75]m, and z=[3,21]m. It shows general improvements by our methods, such as removing occupancy behind parked cars, leading to cleaner occupancy predictions (see top right example). The top left shows a failure case of our method where a moving car produces a drawn-out shadow for our methods, likely resulting from conflicting temporal information. 
}
\label{fig:occ_baselines1}
\end{figure*}

We present additional visualizations of our models and the baselines \ac{bts} \cite{wimbauer2023behind} and PixelNeRF \cite{yu2021pixelnerf} in \autoref{fig:occ_baselines1} and \autoref{fig:occ_baselines2} in the monocular occupancy prediction setting. The examples show overall improvements in our scene reconstruction. Our models produce cleaner edges and show a better holistic scene understanding by reconstructing house facades as straight lines, removing some of the bulges in \ac{bts}. Apart from the overall improvements of our approach compared to \ac{bts}, they also demonstrate some of the limitations of the static scene assumption taken in our model. 
Dynamic objects in the scene can lead to conflicting information in the scene reconstruction of our \ac{mvbts} model, which also affects KDBTS. This results in either drawn-out shadows of dynamic objects (see the first example of \autoref{fig:occ_baselines1}) or our models removing the dynamic object entirely (see the first example of \autoref{fig:occ_baselines2}).

\subsection{Depth Prediction}\label{sup:depth_pred}
We evaluate the depth predictions generated by our models compared to established baselines such as \ac{bts} \cite{wimbauer2023behind} and PixelNeRF \cite{yu2021pixelnerf}, specifically focusing on monocular input.

\begin{figure*}
\centering
\captionsetup{type=figure}
\includegraphics[trim={0cm 0cm 0cm 0cm},clip,width=\textwidth]{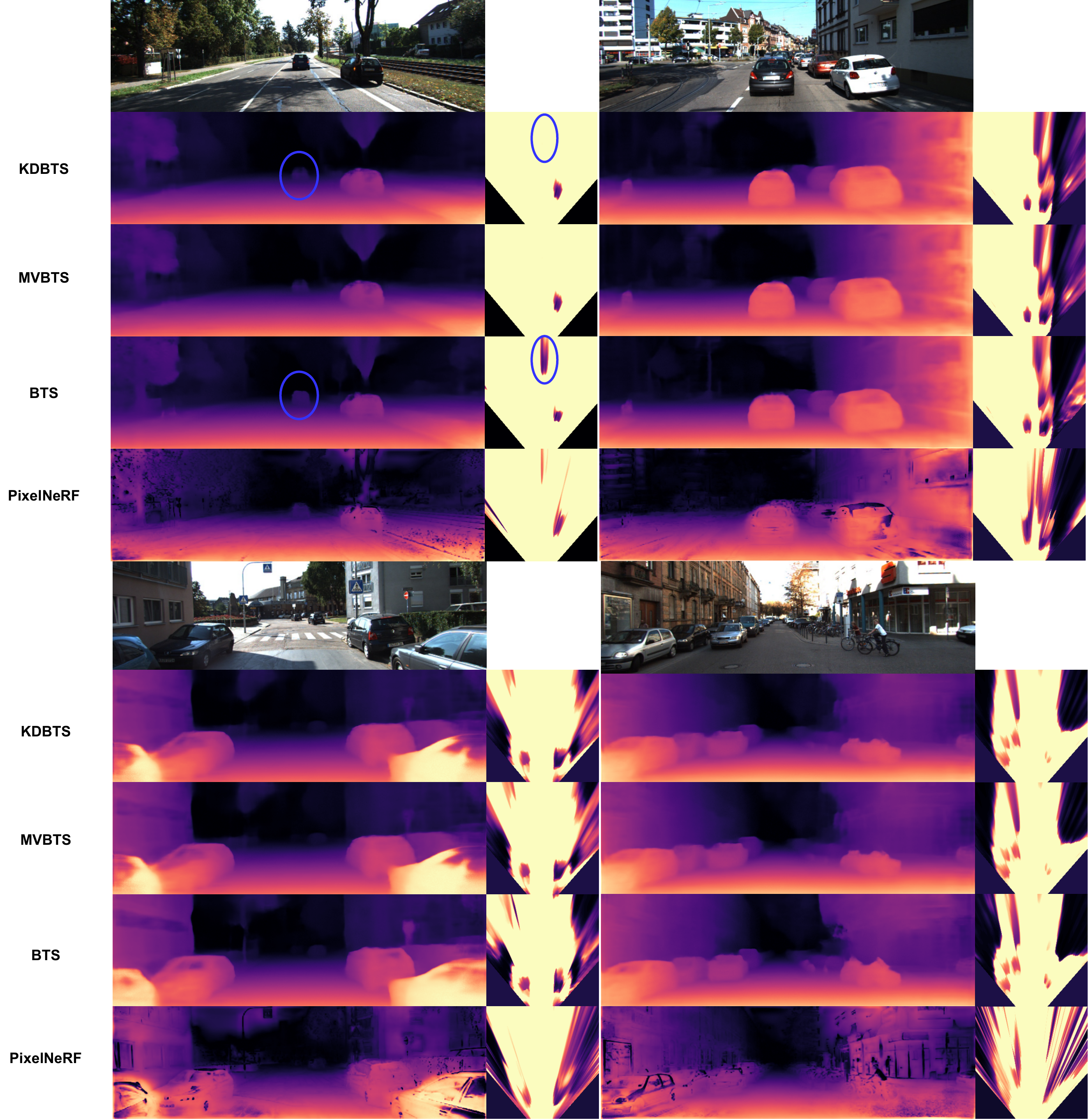}
\captionof{figure}{
\textbf{Qualitative Baseline Results 2.}
		Baselines comparison in both depth- and occupancy estimation. The camera frustum is set up in x=[-9, 9]m, y=[0, 0.75]m, and z=[3,21]m. Our method shows general improvements, such as removing holes in car windows (see lower examples) or predicting the house facades to be in a straight line (see lower right example). It also shows a failure case (top left) where our methods remove a moving car from the scene, likely due to conflicting temporal information.
}
\label{fig:occ_baselines2}
\end{figure*}

Figures \ref{fig:occ_baselines1} and \ref{fig:occ_baselines2} visually present the depth predictions obtained from our models and the aforementioned baselines. Our models demonstrate performance on par with \ac{bts} in terms of overall accuracy. However, a notable difference lies in the prediction of car windows, where our models tend to exhibit fewer holes compared to \ac{bts}. Conversely, our models may show slightly increased blur at the edges of reconstructed objects.

\begin{figure*}
    \captionsetup{type=figure}
    \centering
    \begin{subfigure}[b]{0.48\textwidth}
        \centering
        \includegraphics[trim={0cm 0cm 0cm 0cm},clip,width=\linewidth]{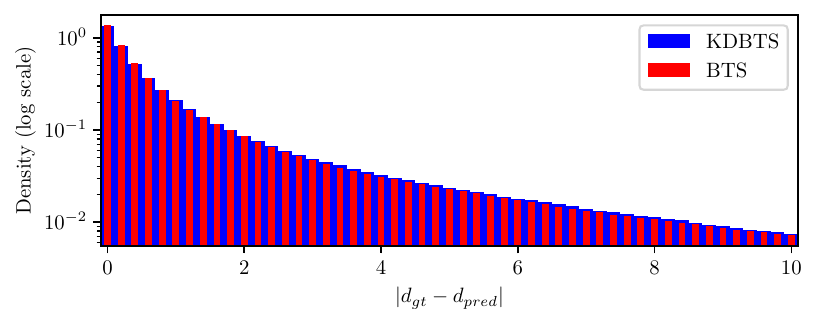}
        \caption{
            absolute error distribution.
        }
        \label{fig:depth_prediction_abs}
    \end{subfigure}
    \hfill
    \begin{subfigure}[b]{0.48\textwidth}
        \centering
        \includegraphics[trim={0cm 0cm 0cm 0cm},clip,width=\linewidth]{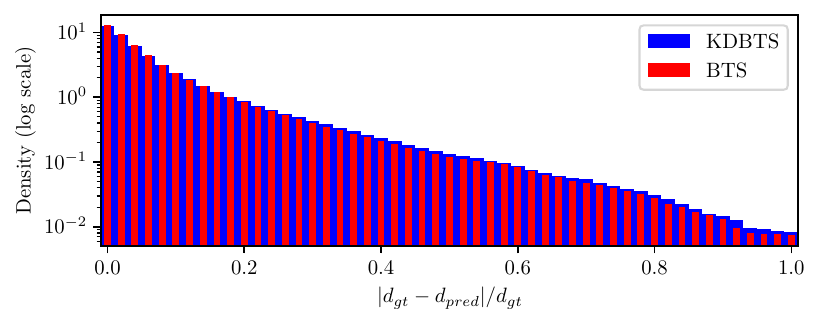}
        \caption{relative error distribution.
        }
        \label{fig:depth_prediction_rel}
    \end{subfigure}
    \caption{
        \textbf{Depth Prediction Error Distributions.} The error distributions are similar for both methods - \textcolor{blue}{\acs{kdbts}} has slightly more large errors than \textcolor{red}{\acs{bts}}. Lower errors for \acs{kdbts} (\textcolor{blue}{blue} dots) and \acs{bts} (\textcolor{red}{red} dots), intensity encodes magnitude. Qualitative examples (see \autoref{fig:Depth_KDMVBTS}) for the depth error on the KITTI test set show that KDTBS often performs worse on dynamic objects (e.g. cars, cyclists).
        }
    \label{fig:depth_prediction_hist}
\end{figure*}

To further analyze the performance differences, we examine the error distribution depicted in \autoref{fig:depth_prediction_hist}. While the majority of errors appear similar between our model (\ac{kdbts}) and \ac{bts}, \ac{kdbts} tends to exhibit more large errors, indicating the violation of static scene assumption.

\begin{figure*}
\centering
\captionsetup{type=figure}
\includegraphics[trim={0cm 0cm 0cm 0cm},clip,width=\textwidth]{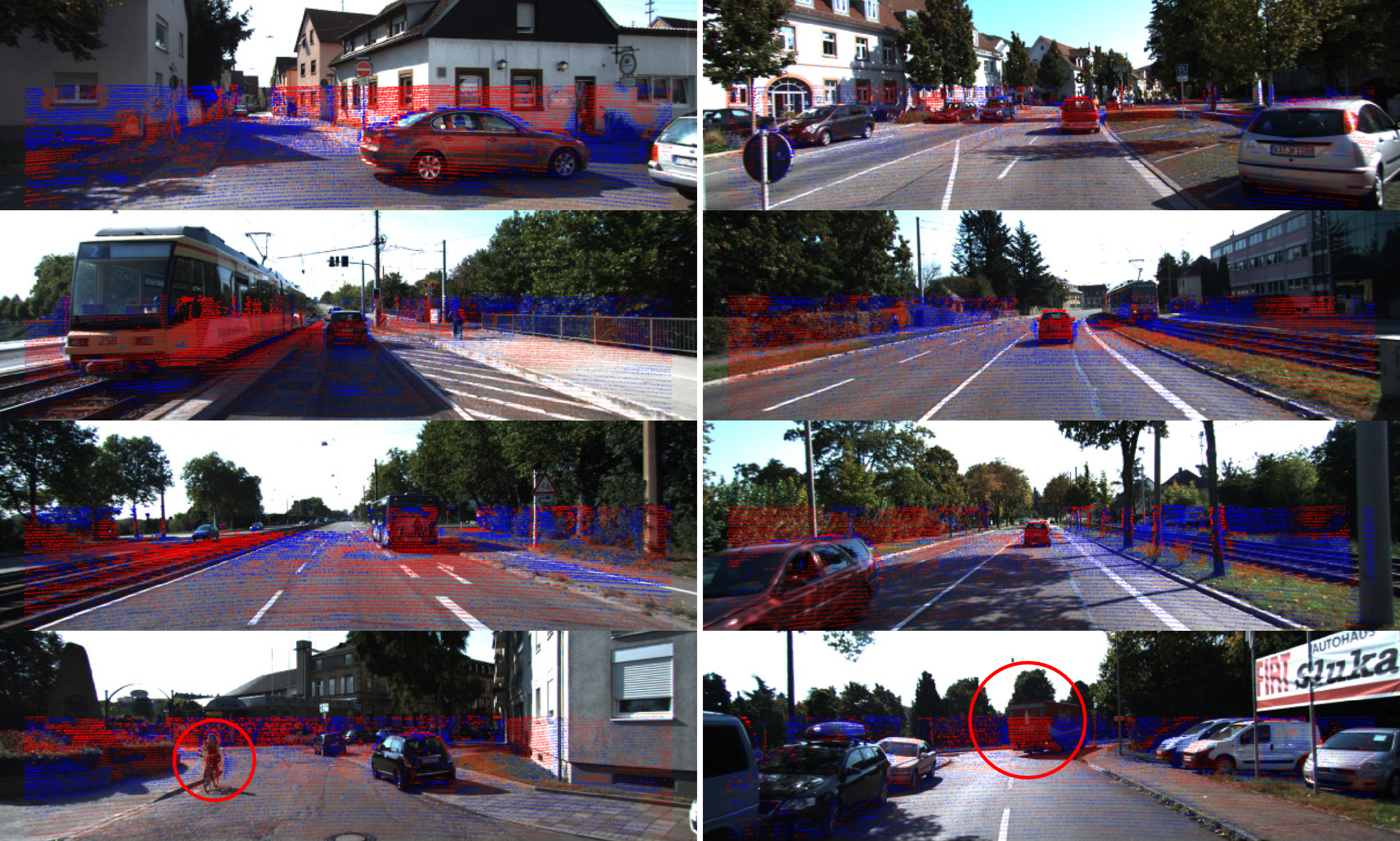}
\captionof{figure}{
\textbf{Depth error comparison between BTS and KDBTS.} \textcolor{blue}{\acs{kdbts}} exhibits slightly more large errors compared to \textcolor{red}{\acs{bts}}. Qualitative examples (bottom) demonstrate depth error on the KITTI test set, with lower errors depicted by \acs{kdbts} (\textcolor{blue}{blue} dots) and \acs{bts} (\textcolor{red}{red} dots), with intensity representing magnitude. Each test image presents the projected scene from the LiDAR ground truth point cloud. The projected LiDAR point cloud is used to calculate the distance error between the prediction and its ground truth. Color differentiation indicates lesser distance errors between \acs{kdbts} (\textcolor{blue}{blue} dots) and \acs{bts} (\textcolor{red}{red} dots). In KDBTS, the model's reconstruction is affected by moving objects, resulting in larger errors typically observed on dynamic objects (e.g., cars, cyclists). Consequently, \textcolor{red}{red} dots are depicted for dynamic objects, signifying lower errors for BTS.
}
\label{fig:Depth_KDMVBTS}
\end{figure*}

In \autoref{fig:Depth_KDMVBTS}, we provide a qualitative comparison of the depth predictions generated by \ac{kdbts} and \ac{bts}. Although both methods perform comparably overall, \ac{kdbts} shows a tendency to perform worse, particularly in scenarios involving dynamic objects. This discrepancy is attributed to violations of the static scene assumption, where the prediction of moving cars may vanish due to conflicting information from multiple time steps (as illustrated in \autoref{fig:occ_baselines2}). These inconsistencies in temporal information impact the reconstruction quality, affecting both depth and occupancy estimates.

\section{Implementation Details}\label{sup:impl_detail}
In the following, we detail the implementation details of our method, including the network architecture and training hyperparameters. For additional details, such as more information about the rendering process and the positional encoding, we refer the reader to \cite{wimbauer2023behind} and its supplementary material.

\subsection{Network Architecture}
For implementation reasons, our network consists of a backbone encoder-decoder network and two decoder networks for both the single-view and multi-view settings, respectively.

\pseudoparagraph{Backbone.} For the backbone, we follow Monodepth2~\cite{godard2019digging} and \ac{bts}~\cite{wimbauer2023behind} such that the reported results stem from the different training setups. It is comprised of a ResNet50 network~\cite{he2015deep_resnet50} with an adjustable channel size of 64. As with \ac{bts}~\cite{wimbauer2023behind}, there is no feature reduction in the upconvolutions of the network.

\pseudoparagraph{Single-View Head.} The single-view decoding network follows \cite{wimbauer2023behind} exactly and is comprised of a layer-connected network with ReLU activation functions and residual connections. The input dimension of the MLP is $103$ ($64$ feature channel size + $39$ positional encoding size) and the network has a hidden dimension of $64$.

\pseudoparagraph{Multi-View Head.} The multi-view decoding network consists of two MLPs with the same architecture as the single-view decoding network. $\text{MLP}_1$, acting as a feature reduction network, has an input dimension of $103$, a hidden dimension of $128$, and an output size of $17$ ($1$ confidence value + $16$ feature channel size). $\text{MLP}_2$ has an input dimension of $16$, and a hidden dimension of $16$. The softmax layer in between the MLPs uses no temperature scaling. 

\pseudoparagraph{Ablation Study Network.} For our ablation study, we also test networks where the MLP dimensions are set to the following for the large network:
\begin{itemize} %
    \item $\text{MLP}_1$ (input: $103$, hidden: $256$, output: $33$)
    \item $\text{MLP}_2$ (input: $32$, hidden: $32$, output: $1$)
\end{itemize}
and to 
\begin{itemize}
    \item $\text{MLP}_1$ (input: $103$, hidden: $128$, output: $2$)
    \item no $\text{MLP}_2$
\end{itemize}
for the small model. The small network does not do feature fusion but rather directly fuses the different frames' density prediction.

\subsection{Training Configuration}
\pseudoparagraph{Hyperparameters}
For training on both KITTI \cite{geiger2013vision} and KITTI-360 \cite{liao2022kitti}, we use the same set of hyperparameters. We use a batch size of $8$ during training. We use the patched-based sampling strategy of \cite{wimbauer2023behind} and sample $32$ random patches of size $8 \times 8$, giving us $2048$ rays in total per batch. The loss weights are set to $\lambda_{SSIM} = 0.85$, $\lambda_{\text{L1}} = 0.15$ following \cite{wimbauer2023behind, godard2019digging} and $\lambda_\text{EAS} = 10^{-3}$ following the code implementation of \cite{wimbauer2023behind}. We use an ADAM optimizer with a learning rate of $\lambda = 10^{-4}$ for the first 120,000 steps and $\lambda = 10^{-5}$ for the rest. We apply the same color augmentation and random flips as \cite{wimbauer2023behind}. For our knowledge distillation, we train with a constant learning rate of $\lambda = 10^{-4}$.

\newpage

\begin{table*}[t]
	\centering
	\footnotesize
	\begin{tabular}{lcccc|ccc|ccc}
		\toprule
		Inference       & dropout & $\text{MLPs}$ & attn. layers              & encode fisheye            & $\text{O}_\text{acc} \uparrow$ & $\text{O}_\text{prec} \uparrow$ & $\text{O}_\text{rec} \uparrow$ & $\text{IE}_\text{acc} \uparrow$ & $\text{IE}_\text{prec} \uparrow$    & $\text{IE}_\text{rec} \uparrow$ \\
		\midrule
		(\textit{Mono}) & 0.5     & middle        & {\cmark} & {\xmark}   & {93.98\%}       & {57.41\%}        & {80.47\%}       & {74.43\%}        & {48.40\%}   & {44.47\%}      \\    %

		(\textit{Mono}) & 0.0     & middle        & {\xmark}   & {\cmark} & {93.81\%}       & {57.07\%}        & {82.60\%}       & {73.03\%}        & {50.55\%}   & {39.54\%}        \\   %

		(\textit{Mono}) & 0.2     & middle        & {\xmark}   & {\cmark} & {94.37\%}       & {59.03\%}        & {80.91\%}       & {76.24\%}        & {51.06\%}   & {{46.28}\%}      \\  %

		(\textit{Mono}) & 0.5     & middle        & {\xmark}   & {\cmark} & {94.71\%}     & {{60.31}\%}        & {83.35\%}       & {{77.89}\%}        & {52.96\%} & {46.02\%}      \\    %

		(\textit{Mono}) & 0.8     & middle        & {\xmark}   & {\cmark} & {{94.78}\%}     & {60.09\%}        & {{85.38}\%}     & {77.76\%}        & {{53.25}\%} & {43.43\%}        \\    %

		(\textit{Mono}) & 0.5     & small         & {\xmark}   & {\xmark}   & {94.51\%}       & {59.62\%}        & {{85.18}\%}     & {76.88\%}        & {52.80\%}   & {40.80\%}        \\    %

		(\textit{Mono}) & 0.5     & large         & {\xmark}   & {\xmark}   & {94.16\%}       & {58.56\%}        & {80.96\%}       & {76.04\%}        & {50.78\%}   & {{46.69}\%}      \\   %
		\midrule
		\midrule
		(\textit{Mono}) & 0.5     & middle        & {\xmark}   & {\xmark}   & {94.76}\%                        & {60.83}\%                         & 84.51\%                        & {78.00}\%                         & {53.69}\%                    & 44.04\%                         \\    %

		\bottomrule
	\end{tabular}
	\caption{\textbf{Ablation Studies in Monocular Inference.} Evaluation of all models in the ablations study in the monocular setting (see \autoref{fig:infer_frame_arrangement} for more details of the frame arrangements). As for the setting in the main paper, our final architecture performs the best with a few exceptions. The model with more dropout performs slightly better in the occupancy estimation task. This is likely due to having fewer frames available during training. Additionally, some of the methods that also encoded fisheye cameras during training as well perform better when it comes to the $\text{IE}_{rec}$ but worse at $\text{IE}_{prec}$. They produce fewer false negatives, but more false positives, meaning less space is predicted as being empty.}
	\label{tab:sup_occ_ablation_mono}
\end{table*}

\begin{table*}[t]
 \centering
	\footnotesize
	\begin{tabular}{lcccc|ccc|ccc}
		\toprule
		Inference       & dropout & $\text{MLPs}$ & attn. layers              & encode fisheye            & $\text{O}_\text{acc} \uparrow$ & $\text{O}_\text{prec} \uparrow$ & $\text{O}_\text{rec} \uparrow$ & $\text{IE}_\text{acc} \uparrow$ & $\text{IE}_\text{prec} \uparrow$    & $\text{IE}_\text{rec} \uparrow$ \\
		\midrule
		(\textit{M}+\textit{T}) & 0.5     & middle        & {\cmark} & {\xmark}   & {93.79\%}       & {58.97\%}        & {77.44\%}       & {72.93\%}        & {45.81\%}   & {{47.91}\%}      \\    %

		(\textit{M}+\textit{T}) & 0.0     & middle        & {\xmark}   & {\cmark} & {93.69\%}       & {57.36\%}        & {81.59\%}       & {73.50\%}        & {45.66\%}   & {41.07\%}        \\   %

		(\textit{M}+\textit{T}) & 0.2     & middle        & {\xmark}   & {\cmark} & {94.42\%}       & {59.45\%}        & {81.76\%}       & {76.45\%}        & {51.44\%}   & {45.94\%}      \\  %

		(\textit{M}+\textit{T}) & 0.5     & middle        & {\xmark}   & {\cmark} & {94.78\%}     & {{60.69}\%}        & {84.37\%}       & {77.93\%}        & {{55.32}\%} & {45.81\%}      \\    %

		(\textit{M}+\textit{T}) & 0.8     & middle        & {\xmark}   & {\cmark} & {{94.90}\%}     & {60.42\%}        & {{86.04}\%}     & {{79.21}\%}        & {53.55\%} & {43.73\%}        \\    %

		(\textit{M}+\textit{T}) & 0.5     & small         & {\xmark}   & {\xmark}   & {94.56\%}       & {59.95\%}        & {{85.98}\%}     & {78.37\%}        & {{55.30}\%}   & {42.65\%}        \\    %

		(\textit{M}+\textit{T}) & 0.5     & large         & {\xmark}   & {\xmark}   & {94.30\%}       & {59.21\%}        & {81.28\%}       & {76.65\%}        & {50.74\%}   & {{47.85}\%}      \\   %
\midrule
		\midrule
		(\textit{M}+\textit{T}) & 0.5     & middle        & {\xmark}   & {\xmark}   & {94.82}\%                        & {61.02}\%                         & 84.83\%                        & {78.73}\%                         & 53.88\%                    & 44.81\%                         \\    %
		\bottomrule
	\end{tabular}
	\caption{\textbf{Ablation Studies in Temporal Monocular Inference.} Using one additional temporal frame shows slight improvements for all methods, with the exception of the attention layer model (see \autoref{fig:infer_frame_arrangement} for more details of the frame setup). Otherwise, the difference between the models is similar to the monocular setting. The model with a higher dropout gains additional performance improvements to our final model.}
	\label{tab:sup_occ_ablation_mono_temporal}
\end{table*}

\begin{table*}[!t]
 \centering
	\footnotesize
	\begin{tabular}{lcccc|ccc|ccc}
		\toprule
		Inference       & dropout & $\text{MLPs}$ & attn. layers              & encode fisheye            & $\text{O}_\text{acc} \uparrow$ & $\text{O}_\text{prec} \uparrow$ & $\text{O}_\text{rec} \uparrow$ & $\text{IE}_\text{acc} \uparrow$ & $\text{IE}_\text{prec} \uparrow$    & $\text{IE}_\text{rec} \uparrow$ \\
		\midrule
		(\textit{S}) & 0.5     & middle        & {\cmark} & {\xmark}   & 93.92\%  &	59.26\%  &	78.57\%  &	73.34\%  &	45.11\%  &	{47.37}\%  \\

		(\textit{S}) & 0.0     & middle        & {\xmark}   & {\cmark} & 93.41\%  &	57.21\%  &	80.23\%  &	72.40\%  &	46.38\%  &	42.14\%  \\  %

		(\textit{S}) & 0.2     & middle        & {\xmark}   & {\cmark} & 94.53\%  &	59.02\%  &	82.04\%  &	76.65\%  &	53.67\%  &	45.97\%  \\  %

		(\textit{S}) & 0.5     & middle        & {\xmark}   & {\cmark} & {94.87}\%  &	{60.09}\%  &	84.58\%  &	{78.37}\%  &	{54.75}\%  &	45.71\%  \\ %

		(\textit{S}) & 0.8     & middle        & {\xmark}   & {\cmark} & {94.89}\%  &	60.07\%  &	{85.64}\%  &	77.98\%  &	54.12\%  &	43.22\%  \\ %

		(\textit{S}) & 0.5     & small         & {\xmark}   & {\xmark}   & 94.56\%  &	59.34\%  &	{86.20}\%  &	77.92\%  &	{54.36}\%  &	42.44\%  \\ %

		(\textit{S}) & 0.5     & large         & {\xmark}   & {\xmark}   &  94.29\%  &	58.86\%  &	81.83\%  &	76.15\%  &	50.90\%  &	{46.15}\%  \\ %
\midrule
		\midrule
		(\textit{S}) & 0.5     & middle        & {\xmark}   & {\xmark}   & 94.84\%  &	{61.12}\%  &	85.28\%  &	{78.62}\%  &	54.02\%  &	44.05\%  \\ %

    \bottomrule
	\end{tabular}
	\caption{\textbf{Ablation Studies in Stereo Inference.} Using stereo frame also shows slight improvements for all methods, with the exception of the attention layer model (see \autoref{fig:infer_frame_arrangement} for more details of the frame setup).}
	\label{tab:sup_occ_ablation_stereo}
 	\vspace{3cm}
\end{table*}

\begin{table*}
\centering
	\footnotesize
	\begin{tabular}{lcccc|ccc|ccc}
		\toprule
		Inference       & dropout & $\text{MLPs}$ & attn. layers              & encode fisheye            & $\text{O}_\text{acc} \uparrow$ & $\text{O}_\text{prec} \uparrow$ & $\text{O}_\text{rec} \uparrow$ & $\text{IE}_\text{acc} \uparrow$ & $\text{IE}_\text{prec} \uparrow$    & $\text{IE}_\text{rec} \uparrow$ \\
		\midrule
		(\textit{S} + \textit{T}) & 0.5     & middle        & {\cmark} & {\xmark}   & 93.23\%  &	{60.77}\%  &	70.69\%  &	69.83\%  &	47.40\%  &	{58.94}\% \\     %

		(\textit{S} + \textit{T}) & 0.0     & middle        & {\xmark}   & {\cmark} & 93.44\%  &	56.85\%  &	79.74\%  &	73.02\%  &	45.54\%  &	43.90\% \\ %

		(\textit{S} + \textit{T}) & 0.2     & middle        & {\xmark}   & {\cmark} & 94.57\%  &	59.85\%  &	83.20\%  &	77.13\%  &	54.41\%  &	45.64\% \\  %

		(\textit{S} + \textit{T}) & 0.5     & middle        & {\xmark}   & {\cmark} & {94.93}\%  &	60.72\%  &	85.47\%  &	79.05\%  &	{55.73}\%  &	46.39\% \\ %

		(\textit{S} + \textit{T}) & 0.8     & middle        & {\xmark}   & {\cmark} & {94.94}\%  &	60.43\%  &	{86.40}\%  &	{79.41}\%  &	{55.35}\%  &	44.80\% \\ %

		(\textit{S} + \textit{T}) & 0.5     & small         & {\xmark}   & {\xmark}   & 94.64\%  &	59.64\%  &	{86.65}\%  &	78.54\%  &	55.04\%  &	43.70\% \\ %

		(\textit{S} + \textit{T}) & 0.5     & large         & {\xmark}   & {\xmark}   &  94.38\%  &	59.88\%  &	82.51\%  &	77.24\%  &	51.66\%  &	{46.90}\% \\ %
\midrule
		\midrule
		(\textit{S} + \textit{T}) & 0.5     & middle        & {\xmark}   & {\xmark}   & 94.91\%  &	{61.73}\%  &	85.78\%  &	{79.47}\%  &	55.08\%  &	45.23\% \\ %

    \bottomrule
	\end{tabular}
	\caption{\textbf{Ablation Studies in Temporal Stereo Inference.} For convenience, we repeat the findings of the main paper here (see \autoref{fig:infer_frame_arrangement} for more details of the frame setup).}
	\label{tab:sup_occ_ablation_stereo_temporal}
\end{table*}

\begin{table*}
\centering
	\footnotesize
	\begin{tabular}{lcccc|ccc|ccc}
		\toprule
		Inference       & dropout & $\text{MLPs}$ & attn. layers              & encode fisheye            & $\text{O}_\text{acc} \uparrow$ & $\text{O}_\text{prec} \uparrow$ & $\text{O}_\text{rec} \uparrow$ & $\text{IE}_\text{acc} \uparrow$ & $\text{IE}_\text{prec} \uparrow$    & $\text{IE}_\text{rec} \uparrow$ \\
		\midrule
		(\textit{S}+\textit{T}+\textit{F}) & 0.5     & middle        & {\cmark} & {\xmark}   & 92.88\%  &	58.51\%  &	71.39\%  &	69.67\%  &	47.31\%  &	{57.28}\%  \\

		(\textit{S}+\textit{T}+\textit{F}) & 0.0     & middle        & {\xmark}   & {\cmark} & 93.22\%  &	55.79\%  &	78.58\%  &	72.36\%  &	44.49\%  &	44.09\%  \\

		(\textit{S}+\textit{T}+\textit{F}) & 0.2     & middle        & {\xmark}   & {\cmark} & 94.54\%  &	58.77\%  &	84.05\%  &	77.19\%  &	55.16\%  &	44.00\%  \\

		(\textit{S}+\textit{T}+\textit{F}) & 0.5     & middle        & {\xmark}   & {\cmark} & {94.90}\%  &	{60.34}\%  &	85.31\%  &	78.84\%  &	{56.08}\%  &	{46.04}\%  \\ %

		(\textit{S}+\textit{T}+\textit{F}) & 0.8     & middle        & {\xmark}   & {\cmark} & {94.91}\%  &	60.23\%  &	{86.22}\%  &	{79.31}\%  &	55.22\%  &	44.50\%  \\ %

		(\textit{S}+\textit{T}+\textit{F}) & 0.5     & small         & {\xmark}   & {\xmark}   & 94.67\%  &	59.17\%  &	{86.65}\%  &	78.54\%  &	54.97\%  &	43.45\%  \\ %

		(\textit{S}+\textit{T}+\textit{F}) & 0.5     & large         & {\xmark}   & {\xmark}   &  94.38\%  &	59.22\%  &	82.68\%  &	77.08\%  &	51.53\%  &	45.90\%  \\ %
\midrule
		\midrule
		(\textit{S}+\textit{T}+\textit{F}) & 0.5     & middle        & {\xmark}   & {\xmark}   & 94.89\%  &	{60.38}\%  &	85.83\%  &	{79.44}\%  &	{55.30}\%  &	44.90\%  \\ %

    \bottomrule
	\end{tabular}
	\caption{\textbf{Ablation Studies in Temporal Stereo Fisheye Inference.} Using the fisheye cameras for inference does not give large improvements for all methods. This shows that a lot of the scene information can already be captured in the pinhole frames alone (see \autoref{fig:infer_frame_arrangement} for more details of the frame setup).}
	\label{tab:sup_occ_ablation_stereo_temporal_fisheye}
\end{table*}

\newpage

\newpage

\newpage

\end{document}